\documentclass[10pt]{wlscirep}
\usepackage[utf8]{inputenc}
\usepackage[T1]{fontenc}
\usepackage{graphicx}
\usepackage{multirow}
\usepackage{amsmath}
\usepackage{longtable}
\usepackage{booktabs}
\title{Enhancing Skin Cancer Diagnosis (SCD) Using Late Discrete Wavelet Transform (DWT) and New Swarm-Based Optimizers}

\author[1]{Ramin Mousa}
\author[2,+]{Saeed Chamani}
\author[3,+,*]{Mohammad Morsali}
\author[3,]{Mohammad  Kazzazi}
\author[3]{Parsa Hatami}
\author[4]{Soroush Sarabi}
\affil[1]{University of Zanjan, Department of Computer Engineering, Zanjan, Iran}
\affil[2]{Iran University of Science and Technology, Department of Biomedical Engineering, Tehran, Iran}
\affil[3]{Sharif University of Technology, Department of Electrical Engineering, Tehran, Iran}
\affil[4]{Raderon AI Lab, British Columbia, Canada}

\affil[*]{mohammad.morseli@ee.sharif.edu, Corresponding Author}

\affil[+]{these authors contributed equally to this work}

\begin{abstract}
Skin cancer (SC) stands out as one of the most life-threatening forms of cancer, with its danger amplified if not diagnosed and treated promptly. Early intervention is critical, as it allows for more effective treatment approaches, potentially halting the disease's progression and improving patient outcomes. In recent years, Deep Learning (DL) has emerged as a powerful tool in the early detection and skin cancer diagnosis (SCD). Although the DL seems promising for the diagnosis of skin cancer, still ample scope exists for improving model efficiency and accuracy. This paper proposes a novel approach to skin cancer detection, utilizing optimization techniques in conjunction with pre-trained networks and wavelet transformations. First, normalized images will undergo pre-trained networks such as Densenet-121, Inception, Xception, and MobileNet to extract hierarchical features from input images. After feature extraction, the feature maps are passed through a Discrete Wavelet Transform (DWT) layer to capture low and high-frequency components. Then the self-attention module is integrated to learn global dependencies between features and focus on the most relevant parts of the feature maps. The number of neurons and optimization of the weight vectors are performed using three new swarm-based optimization techniques, such as Modified Gorilla Troops Optimizer (MGTO), Improved Gray Wolf Optimization (IGWO), and Fox optimization algorithm, to improve the efficacy of the model and diagnostic accuracy. Evaluation results demonstrate that optimizing weight vectors using optimization algorithms can enhance diagnostic accuracy and make it a highly effective approach for SCD. The proposed method demonstrates substantial improvements in accuracy, achieving top rates of 98.11\% with the MobileNet + Wavelet + FOX and DenseNet + Wavelet + Fox combination on the ISIC-2016 dataset and 97.95\% with the Inception + Wavelet + MGTO combination on the ISIC-2017 dataset, which improves accuracy by at least 1\% compared to other methods.

\end{abstract}
\begin{document}

\flushbottom
\maketitle
%
%
\thispagestyle{empty}


\section*{Introduction}

SCD is one of the dangerous diseases that is treatable with early diagnosis but still claims the lives of many people annually. In 2022, 331,722 people were diagnosed with skin cancer, and 58,667 of them lost their lives due to the disease, and these numbers are increasing every year\cite{wcrf_skin_cancer_statistics}.\\
Early diagnosis plays a crucial role in the treatment process and the patient's chance of survival. It can significantly increase the likelihood of the disease being operable, allow treatment methods to be applied more effectively, and prevent the disease from progressing\cite{Siegel2024Cancer}.\\
Skin cancer symptoms often appear as skin lesions that look very similar to harmless ones and can only be recognized by dermatologists. Because of this, many people with the disease ignore these skin changes and do nothing until the cancer has advanced. This delay in diagnosis can slow down treatment, make it less effective, and allow the disease to spread to other body parts. A significant reason for this delay is the difficulty of current diagnostic methods, like sampling.\\
Machine learning and deep learning can play a significant role in skin cancer detection by providing automated detection. The most critical issue in using learning algorithms for disease detection is their accuracy. Therefore, improving the accuracy of these algorithms is of utmost importance. As we know CNN models perform very well in classification problems. Thus, we use CNN as our base model. In this article, we increased the model's accuracy compared to previous models. The key innovations we employed in this paper to enhance accuracy are the use of the swarm-based optimizers and the integration of the Wavelet Transform.\\
In this article, we aimed to significantly increase the accuracy of the model used for disease diagnosis. First, we preprocessed the data in the dataset and resolved the data imbalance problem, and then the pictures were ready for giving as the input to the model. To improve the model's accuracy, we employed innovations that can be broadly divided into the following two stages: \\

\begin{itemize}
    \item The images are fed into four pre-trained models. In the feature extraction section, we utilized the wavelet transform to extract suitable features for training the data. The output is then passed through a self-attention mechanism. Since the classification task is binary, the output is passed through a sigmoid function.

    \item After completing all these stages, the network is fed into an ANN, where three advanced and modern optimizers (IGWO, FOX, MGTO) are used to update and modify the model's parameters, and their performance is evaluated. These optimizers update the parameters optimally, which significantly improves the model's accuracy and overall performance.
\end{itemize}

In previous studies on SCD, the methods, tools, and optimizers mentioned above were not utilized this way. The combination of using wavelet transforms for feature extraction, evaluating multiple new optimizers, and analyzing four different models has led to significant improvements in our results compared to prior studies in this field. This approach could be an important step in advancing deep models for diagnosing various diseases based on medical images.

\section*{Literature review}
In\cite{Du_2023}, first, FairDisCo, a deep neural network (DNN) with contrastive learning, was proposed. This model includes
a network branch that reduces sensitive details from model representations and another contrastive branch to improve
representation learning and thus improve SCD diagnosis accuracy. Secondly, introduced MDViT, a multi-domain Vision
Transformer (VIT) with domain adapters. MDViT aims to handle model data hunger and negative knowledge transfer (NKT),
which can reduce model performance on domains with inter-domain heterogeneity. MDViT also uses mutual knowledge
distillation to improve representation learning across domains. 

In\cite{Ojha2023}, first, noisy samples are removed from the data using an Adaptive Median Filter (AMF). Then, features are extracted from segmented data using Kernel Principal Component Analysis (KPCA). Finally, an Augmented May Fly optimized with the K-Nearest Neighbors (AMFO-KNN) technique improves SCD classification.In\cite{Simon_Vicent_Addah_Bamutura_Atwiine_Nanjebe_Mukama_2023}, compared commonly used pre-trained models for detecting SCD using the same dataset to determine which one has the highest accuracy. This research was done to prevent biased results that can happen when using only one model and dataset in various methods by introducing which model is best to train the SCD datasets.In\cite{HamnaAyesha_AhmadNaeem_AliHaiderKhan_KamranAbid_NaeemAslam_2023}, a new combined model, Multi-Model Fusion for Skin Cancer Detection (MMF-SCD), has been introduced. This model uses pre-trained VGG-16, VGG-19, and ResNet-50 models to extract features accurately. After feature extraction, feature optimization is performed, and the model predicts diseases. This method is proposed to deal with the accuracy limit in SCD classification that exists in several methods.

In\cite{10194499}, presented a new multi-scale cross-attention method called M-VAN Unet for skin lesion segmentation network, based on the Visual Attention Network (VAN). This method handles the challenge of learning both local and global features. Unlike Transformers, which were initially developed for natural language processing (NLP) and may destroy the multidimensional nature of images, M-VAN Unet preserves the 2D structure of the image and can capture low-level features more effectively. In addition, they introduced two attention mechanisms: MSC-Attention for multi-scale channel attention and LKA-Cross-Attention, a cross-attention mechanism based on large kernel attention (LKA). These mechanisms enable the capture of low-level features and simplify global information interaction.

In\cite{cancers15072146}, developed the Optimal Multi-Attention Fusion Convolutional Neural Network-based Skin Cancer Diagnosis (MAFCNN-SCD) technique for detecting skin cancer. This technique applies the MAFNet method as a feature extractor, using the Henry Gas Solubility Optimization (HGSO) algorithm as a hyperparameter optimizer. Finally, the Deep Belief Network (DBN) method is used to detect skin cancer disease.In\cite{diagnostics13050912}, proposed a collaborative learning deep convolutional neural networks (CL-DCNN) model based on the teacher-student learning method for SCD segmentation and classification and provided a self-training method to generate high-quality pseudo-labels. They also employed class activation maps to improve the segmentation network's location ability.

In\cite{diagnostics13081460}, developed a segmentation model called the Mobile Anti-Aliasing Attention U-Net model (MAAU). This model includes encoder and decoder paths. The encoder uses an anti-aliasing layer and convolutional blocks to decrease the spatial resolution of input images while preventing shift equivariance. The decoder uses an attention block and decoder module to capture important features in each channel. In\cite{jennifer2023sickle}, some transfer learning models have been implemented for SCD classification. Advanced image augmentation techniques were used, and these models were combined with a Random Forest classifier and a
support vector machine classifier to see the contribution of different elements to the model's accuracy.

In\cite{son2021ai}, skin images were first preprocessed by decomposing these into their hemoglobin and melanin constituents. Then, the resultant U-Net architecture for segmentation was used, and then a convex hull algorithm was applied to cluster the areas of interest that were inputted into the EfficientNet model for classification. In\cite{oliveira2023multi}, the architectural model of the VENet regarding multitask learning is employed for classifying and segmenting CVDs simultaneously, as in U-Net. It preprocesses the input images first, followed by segmentation, and further classifies the severity levels based on the Clinical Etiologic Anatomic Pathophysiologic (CEAP) protocol. 

In\cite{liu2023one}, a single fiber OCT system was used to perform invivo skin imaging to classify normal and abnormal skin tissues, especially in Mohs micrographic surgery. Original skin layer segmentation was done using the U-Net CNN model, automatically extracting features, after which the detection of these anomalies is conducted through a one-class SVM without demanding large labeled datasets regarding abnormal tissues. In\cite{moldovanu2023refining}, it uses geometric features of superpixels to improve the detection of melanoma and nevi. After preprocessing, the method first applies an improved Simple Linear Iterative Clustering (iSLIC) algorithm to segment dermoscopy images and remove unnecessary background superpixels. It extracts the geometric features used for skin lesion classification after segmentation, such as perimeter, area, and eccentricity.

In\cite{innani2023generative}, a GAN-based framework known as EGAN was implemented for automatic skin lesion segmentation. The overall working diagram for the method includes preprocessing first and then segmentation with the help of an encoder-decoder architecture generator. The proposed framework couples the discriminator along with the PatchGAN architecture to discriminate between real versus synthetic segmentation masks. Besides, a morphology-based smoothing loss function enhances the smoothness and accuracy of the lesion boundaries.

In\cite{lai2023skin}, discussing that various methods of machine learning and optimization have been proposed for Skin Cancer Diagnosis (SCD). After preprocessing, SVM is used to classify the data. Further, some techniques take on image histogram analysis and color characteristics to identify the regions of the lesion. It has also been combined with ANN to optimize image segmentation or weight adjustment. Furthermore, CNN and SVM were employed for feature extraction and classification. Improved Gray Wolf Optimization (IGWO) algorithm have been utilized in the setting of hyperparameters for CNNs to improve the accuracy of classification.

In\cite{singh2024shifting}, concerning the limited labeled medical data, the authors propose a pipeline called S4MI for self-supervision and semi-supervision. This approach uses several self-supervised learning methods, such as DINO and CASS, for uninhibited feature extraction. Meanwhile, semi-supervised learning, especially cross-teaching of CNNs and Transformers, is adopted to further improve the performance of segmentations at a reduced usage of 50\% of labeled data. In\cite{ali2023iomt}, conditional Generative Adversarial Network (cGAN) was used to segment melanoma lesions. This approach first preprocesses the dermoscopic images, and then the segmentation is done using the cGAN model, which maps a dermoscopic image to the segmented lesion image using the network. It consists of an adversarial process by which the generator learns to produce valid segmentation masks while the discriminator identifies if a mask is real or generated.

 In\cite{akram2023segmentation}, a hybrid deep-learning approach has been proposed for the segmentation and classification of skin lesions. It combines two advanced deep learning neural networks: Mask Region-based Convolutional Neural Network (MRCNN) to perform semantic segmentation and ResNet50 to classify the lesions.In\cite{articleIGWO},  The Improved Grey Wolf Optimizer (I-GWO) was utilized to optimize the hyperparameters of different Pretrained Deep Learning Networks (PDLNs) to maximize results for binary image classification of skin cancer. The researchers determined the maximum number of I-GWO iterations and the number of search agents. The lower and upper limits for the required parameters are then determined to get the best hyperparameters’ values and the highest accuracy.
 
In\cite{can}, a new technique is introduced to improve the segmentation process by combining expert-generated and computer-generated labels. The method uses a trained model to generate labels for new data. The results suggest that this method could improve the accuracy of SCD tools.In\cite{lama2023skin}, a new approach for skin lesion segmentation in dermoscopic images was developed using a deep learning model to handle noisy data challenges in segmentation. For this time, the authors proposed a deep learning model for segmentation based on the modified U-Net architecture but containing a pre-trained EfficientNet as the encoder, including squeeze and excitation residual blocks in the decoder portion to boost the feature representation.

In\cite{ali2024novel}, skin cancer detection was proposed with a new fully convolutional encoder-decoder network (FCEDN) combined with a hyper-parameter optimized sparrow search algorithm (SpaSA). After preprocessing, image segmentation is affected by FCEDN with trainable encoder and decoder layers. Skin lesions are further classified into benign and malignant classes by an adaptive convolutional neural network. SpaSA is applied to optimize hyperparameters in FCEDN to extract high-level global features for effective segmentation.

\section*{Methodology}
Our proposed method introduces a unique structure by combining pre-trained Convolutional Neural Networks (CNNs), discrete wavelet transforms (DWT), self-attention mechanisms, and swarm-based optimization techniques to optimize performance, as shown in Figure 1. 
This approach improves the diagnostic accuracy and generalizability of DL models for SCD compared to previous methods.
First, we preprocessed our raw dataset(ISIC-2016 and ISIC-2017) to provide compatibility with the proposed models.
We use image augmentation to train images, using rotation, flipping, zooming, and translation to improve the model's generalization capabilities and class imbalance issues. Then, we normalize images to 200×200 and normalize them to the range [0,1] to standardize the input data. At last, melanoma and non-melanoma labels are mapped into binary labels for classification.

The proposed method's backbone uses pre-trained CNN architectures, such as DenseNet-121, Xception, MobileNet, and Inception, to extract hierarchical features from input images. After feature extraction, the feature maps are passed through a Discrete Wavelet Transform (DWT) layer. DWT decomposes feature maps into four sub-bands (LL, LH, HL, and HH) to capture low- and high-frequency components. Low-frequency (LL) sub-bands emphasize global structures, while high-frequency (LH, HL, HH) sub-bands preserve critical details like edges and textures essential to detect melanoma disease. Finally, sub-band features are concatenated to form a representation of subsequent layers.
After the DWT part, the self-attention module is integrated to learn global dependencies between features and focus on the most relevant parts of the feature maps. Eventually, the wavelet-transformed and attention-refined features are flattened and passed through a dense classification head. Feature vectors are refined through multiple routing iterations, and a sigmoid-activated dense layer outputs the probability of melanoma presence.

To optimize model performance, three advanced swarm-based optimization algorithms, Improved Grey Wolf Optimizer (IGWO), Modified Gorilla Troops Optimizer (MGTO), and Fox Optimizer (FOX), are employed with a simple artificial neural network(ANN). These algorithms optimize the filters' size, kernel size, learning rate, L2 and L1 regularization factors, batch size, and number of epochs.

The optimized model is trained using a custom learning rate scheduler function, which dynamically adjusts learning rates to balance convergence and stability. Binary cross-entropy loss is used to minimize the error of binary classification. 
The proposed method was implemented in Python with TensorFlow and Keras frameworks and Mealpy Library\cite{van2023mealpy,van2023groundwater}. Experiments with the following configurations were conducted on the ISIC-2016 and ISIC-2017 datasets. 
Baseline models include DenseNet, MobileNet, Xception, and Inception without wavelet or optimizers and enhanced models such as Wavelet-transform-enhanced networks with swarm-based optimizers.
\begin{figure}
    \centering

        \includegraphics[width=18cm, height=18cm]{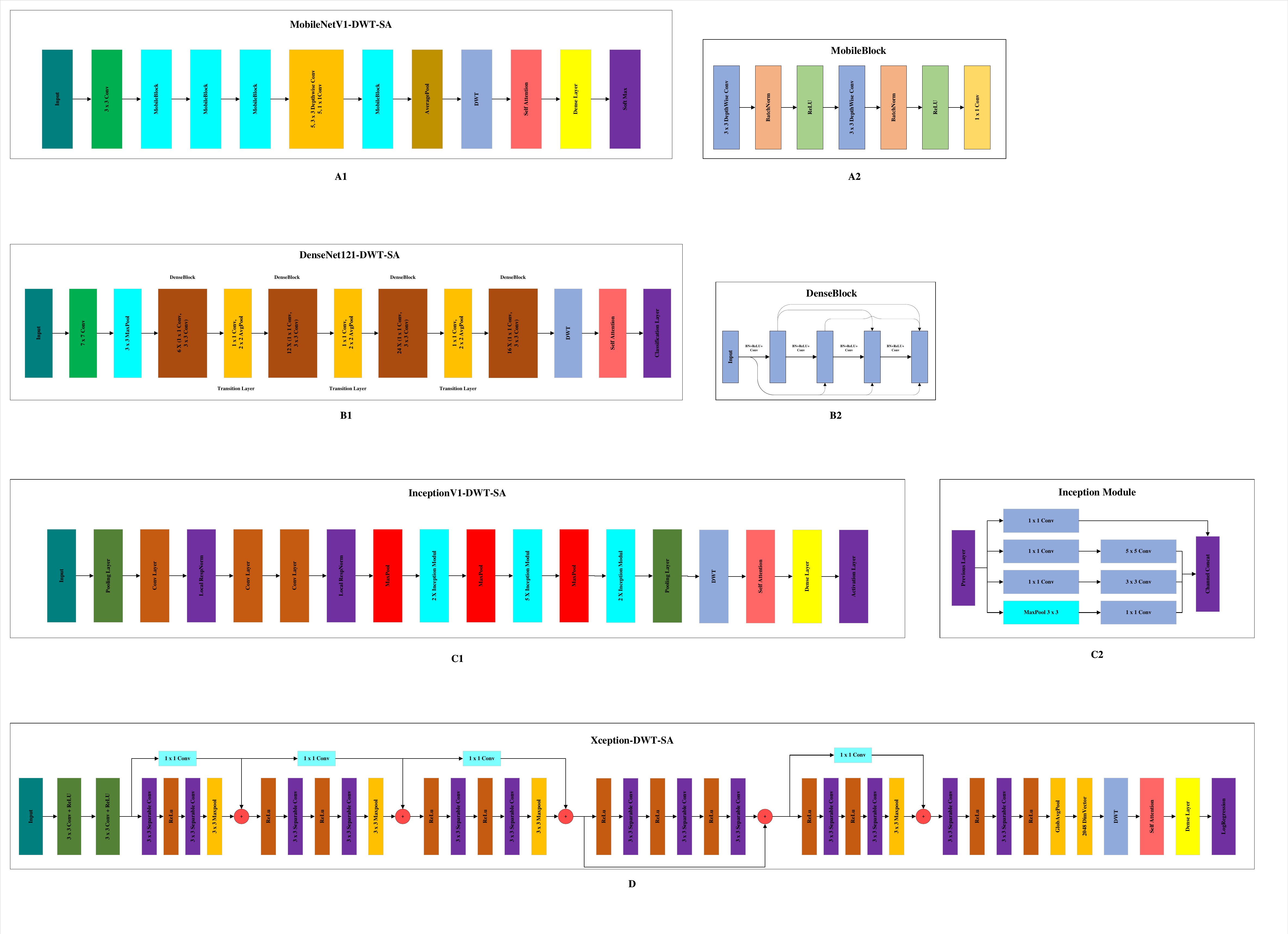}\\

    \caption{Structure of  Proposed Method(The Alphabetical Abbreviations Stands for A1:MobileNetV1-DWT-SA Block, A2: Mobile Net Block, B1: DenseNet121-DWT-SA Block, B2: Dense Net Block, C1: InceptionNetV1-DWT-SA Block, B2: Inception Net Block, D: XceptionNet-DWT-SA Block)}
    \label{fig:architectures} 
\end{figure}

\subsection*{Wavelet Transform}
In image analysis, especially medical images, most images consist of areas with gradually changing pixel intensity. However, these areas are interrupted by some abrupt changes in pixels, often seen in the sharp edges or corners. The interesting point is that critical information is usually present in these abrupt areas because they represent essential features like edges, texture boundaries, or contrasts. Traditional methods like the Fourier transform are unsuitable for analyzing these areas because the Fourier transform breaks down an image into a sum of sine waves, which are not localized in time or space. Therefore, it struggles to display abrupt areas like edges or sharp contrasts accurately\cite{Burac2021}.

The Wavelets can be examined in both time and frequency domains, serve as waves of bounded duration, and exist in different sizes and shapes, with the most important advantage being that there is a wide variety of wavelets to choose from. Wavelet transform is mainly based on two basic standard operations: scaling and shifting. Scaling refers to the stretching or compressing signal in the time domain, which is inversely proportional to frequency. Therefore, shrunken wavelets correspond to higher frequencies for smaller-scale factors; for larger-scale factors, the stretched wavelets correspond to lower frequencies. Stretched wavelets capture the slow variations in an image, while shrunken wavelets capture the abrupt changes. Another critical factor is shifting, which will delay or advance the signal against its features. By looking into the signals at an intermediate scale between octaves, continuous wavelet transforms become handy in time-frequency analysis or filtration of frequency components localized at a specific region of interest. However, discrete wavelet transforms have lower coefficient redundancy because of translation by integer multiples, thus becoming ideal for de-noising and compressing signals and images. Discrete wavelet transforms use the multilevel wavelet decomposition that splits the signal into approximation-low pass and details-high pass subbands, where appropriate threshold techniques might be performed for further analysis\cite{article_wavelet,Chen2022}.

In this architecture, we use late wavelet transformers to extract data more efficiently. The wavelet transformation decomposition comprises four components, one being a low frequency and the other three being high frequencies. Once the CNN processes the input data and produces latent feature maps, the wavelet transform decomposes these into multiple sub-bands representing different frequency components. This decomposition lets the model separate fine details (high-frequency components) from larger structures (low-frequency components). It will improve the model's ability to detect global and local patterns. This is particularly valuable for handling intricate patterns, such as textures or objects appearing at different scales within the input image\cite{Fujieda2018}. Utilizing the wavelet transform after CNN layers enables the system to discern finer details that the convolutional filters alone may not fully resolve, particularly in challenging tasks like image recognition or texture analysis, which is precisely what we want to do\cite{Fujieda2017}. 

\subsection*{Pre-traind Networks}

Inception(also known as GoogleNet) was introduced by Szegedy et al. in 2015\cite{DBLP:journals/corr/SzegedyLJSRAEVR14}. Inception achieved a high accuracy of 93.3\% and won the Top-5 at the 2014 ILSVRC challenge\cite{ILSVRC15}. The architecture consists of 22 layers and features a building block called the “inception module.” 
These modules include Network-in-Network layers, large and small convolutional layers, and a pooling layer. This design helps to stack and configure layers flexibly within CNNs. Additionally,  this architecture handles computations in parallel to reduce the dimensionality by using 1 × 1 convolutions.

Xception(also known as Extreme Inception ), developed as an extension of the Inception architecture, is a CNN that relies exclusively on depthwise separable convolution layers\cite{DBLP:journals/corr/Chollet16a}. This design is based on the idea that cross-channel and spatial correlations within feature maps can be fully separated. The 36 convolutional layers are structured into 14 modules, all of which have linear residual connections around them, except for the first and last modules, allowing for a more efficient and streamlined processing flow.

DenseNet-121\cite{huang2017densely}, introduced by Huang et al. (2017), further utilized the concept of dense connectivity; meaning each layer was connected to all previous and subsequent layers. This allows a reduction in vanishing gradient issues, propels the process of feature propagation, and hence makes training more viable. The growth rate at which a few feature maps are added at each layer with dense blocks and transition layers implement batch normalization, ReLU, 1×1 convolutions, and pooling. This is how competitive accuracies are achieved with DenseNet-121 while still being resource-efficient with fewer parameters for large-scale tasks.

MobileNet\cite{howard2017mobilenets},introduced by Howard et al. (2017), is an efficient neural network designed for mobile and embedded applications introduced by Howard et al. in 2017. It achieves high performance with low computational cost and model size by utilizing depthwise separable convolutions. Two key hyperparameters, the width multiplier and resolution multiplier, can be tuned by changing respectively the channel counts and input resolution against different constraints. MobileNet is also more versatile, suitable for tasks ranging from object detection to fine-grained classification and even face attribute recognition that are highly in demand for on-device intelligence.

\subsection*{Optimization Algorithms}
\subsubsection*{FOX} 

The FOX optimization algorithm is based on the hunting behaviors of foxes in the wild.The algorithm can mimic the fox’s strategy of randomly searching for prey in snowy conditions by relying on its ability to hear the ultrasounds radiated by the prey. The fox estimates the distance to the prey by analyzing the sound and calculates the precise jump needed to catch the prey\cite{Mohammed2022FOXAF}. 
 
The working steps of the FOX optimization algorithm are as follows : 

1. During snow coverage, locating prey becomes notably challenging. The red fox searches the ground randomly to pursue prey.

2. As a result of the search, the red fox locates its prey by detecting its ultrasound sound and then begins approaching the target; this process takes some time.

3. The red fox can determine the distance to its prey by calculating the sound of the prey and the time difference.

4. The red fox jumps toward its prey after estimating the distance.

5. Random walks are taken based on the shortest time and optimal position.

A study found that foxes prefer to jump northeast due to magnetic alignment, with an 82\% success rate in catching prey. If they jump in the opposite direction, the success rate is only 18\%\cite{Sharma2023,Bas2023}.

The algorithm balances exploration and exploitation using a random variable with a 50\% probability for dividing iterations. The algorithm should avoid getting stuck in local optima by balancing two phases. Also, the variable "a" is used to decrease the search performance according to BestX. Its value decreases after each iteration to improve the agent's pursuit of the prey. Additionally, a condition is provided to update the position, and the fitness value affects the search agents to avoid local optima. If the new position does not change, the exploration phase is deactivated to activate other phases.

\textbf{Exploitation Phase:} If the random variable \(p\) is greater than 0.18, we must locate the red fox's new position. So, we need to compute the jumping distance (\(Jump_{it}\)), the distance between the fox and the prey (\(Dist\_Fox\_Prey_{it}\)), and the time it takes for sound to travel (\(Dist\_S\_T_{it}\)). Also, We have the sound travel time \(Time\_S\_T_{it}\), which is a random number ranging from 0 to 1. We can calculate the value of (\(Dist\_S\_T_{it}\)) as below :
\begin{equation}
     Dist_{S_{it}} = Sp_S \times Time\_S\_T_{it}
\end{equation}
   
The number of iterations, denoted by \(it\), ranges from 1 to 500. Also, We have another formula to calculate \(Sp_S\) based on the best location. The best and random agents are referred to as \(BestPosition_{it}\) and \(Time\_S\_T_{it}\). The equations for determining \(Sp_S\) and \(Dist\_Fox\_Prey_{it}\) is as follows:
\begin{equation}
    Sp_S = \frac{BestPosition_{it}}{Time\_S\_T_{it}}
\end{equation}
\begin{equation}
    Dist\_Fox\_Prey_{it} = Dist\_S\_T_{it} \times 0.5
\end{equation}

Once the fox finds the distance from its prey, it must jump to catch it and move to a new location. The equation below calculates the jump height:

\begin{equation}
    Jump_{it} = 0.5 \times 9.81 \times t^2
\end{equation}
\begin{equation}
    tt = \frac{\text{sum}(\text{Time}_{\text{St}}(i,:))}{\text{dimension}}, \quad \text{MinT} = \text{Min}(tt)
\end{equation}
The average time for sound to travel, represented by \(t\), is squared due to the up-and-down movement, and 9.81 represents the acceleration due to gravity. By dividing the sum of the \(Time\_S\_T_{it}\) to dimensions, the value of time transition \(tt\) can be calculated. To determine the mean time \(t\), we divide \(tt\) by 2. The jump happens up and down two times, so we multiply the gravity and meantime by 0.5. After that, we multiply the jump value by \(c_1\) and \(Dist\_Fox\_Prey_{it}\). Each time the fox jumps northeast, the variable \(c_1\) is in the range of [0, 0.18]. The fox's new position is calculated using the below formula when \(p\) is greater than 0.18 by (6) and when it is smaller than 0.18 by (7).
\begin{equation}
    X_{i+1} = Dist\_Fox\_Prey_{it} \times Jump_{it} \times c_1
\end{equation}
\begin{equation}
    X_{i+1} = Dist\_Fox\_Prey_{it} \times Jump_{it} \times c_2
\end{equation}
  \textbf{Exploration phase:} The red fox performs a random walk using its best position at this stage, with a minimum time variable \textit{MinT} and a control variable. It is calculated by the following equation.
\begin{equation}
    a = 2 \times \left( it - \left(\frac{1}{Max_{it}}\right) \right) 
\end{equation}
Where \(Max_{it}\) is the maximum iteration, the novel position of the red fox in the search space is calculated by Equation (9).
\begin{equation}
   X_{(i+1)} = BestX_{t} \times \text{rand}(1, \text{dimension}) \times MinT \times a 
\end{equation}

\subsubsection*{Improved Gray Wolf Optimizer(IGWO)}
Gray Wolf Optimizer (GWO)\cite{NADIMISHAHRAKI2021113917} is a nature-inspired meta-heuristic algorithm that mathematically models the social hierarchy and hunting mechanism of grey wolves. In GWO, the alpha, beta, and delta wolves guide the optimization process by directing other wolves, known as omega, to update their positions. This algorithm imitates the major phases of hunting by a grey wolf: searching for a prey (exploration), encircling the prey (transition between exploration and exploitation), and attacking the prey (exploitation). This approach enables GWO to explore the search space for optimal solutions and converge toward global optima by balancing exploration and exploitation. Despite its simplicity and efficiency, the standard GWO generally performs not so well when dealing with complex optimizations, especially in cases of large searching spaces or when fast convergence speed is required.

The Improved Grey Wolf Optimizer has been developed to respond to these limitations by making some adjustments to improve the adaptability of the algorithm in solving structural optimization problems. In IGWO, dynamically adjustable parameters are considered, which control the exploration and exploitation phases more effectively compared to the original GWO version. While the exploration phase has the duty of scanning widely over the solution space to avoid local optima, the refinement of the search around the best-found solutions is left to the exploration phase. The incorporation of exponential decay functions for the control parameters allows the gradual shifting of IGWO from exploration to the process of exploitation, hence making a focused and efficient search in later iterations. These changes reduce the computational effort and converge faster to provide a more precise solution for complex structural design problems.

The IGWO method differs from the GWO because it uses three functions to enhance exploration and exploitation properties. The parameter \(a\), which is used for alpha, beta, and delta, exponentially decreases over time, as shown by the following equations\cite{Zakian2021}:
\begin{equation}
    a_{\alpha}(i) = a_{\text{max}} \exp \left( \frac{i}{i_{\text{max}}} \eta_{\alpha} \ln \left( \frac{a_{\text{min}}}{a_{\text{max}}} \right) \right), 
\end{equation}
\begin{equation}
    a_{\delta}(i) = a_{\text{max}} \exp \left( \frac{i}{i_{\text{max}}} \eta_{\delta} \ln \left( \frac{a_{\text{min}}}{a_{\text{max}}} \right) \right),
\end{equation}
\begin{equation}
    a_{\beta}(i) = \frac{a_{\alpha}(i) + a_{\delta}(i)}{2}.
\end{equation}

Where \(a_{\text{max}}, a_{\text{min}}, \eta_{\alpha}, \eta_{\delta}, i_{\text{max}}\), and \(i\) indicate the upper bound of \(a\), lower bound of \(a\), growth factor for alpha, growth factor for delta, the maximum number of iterations, and the current number of iterations. 

These functions represent the wolf hierarchy: alphas lead with the lowest search intensity, deltas have the highest, and betas are in between. As a result, \(a_{\delta}\) is the largest, \(a_{\alpha}\) is the smallest, and \(a_{\beta}\) is the average of the two.

 Only two parameters \(a_{\text{max}}\) and \(a_{\text{min}}\) are required to be tuned to achieve faster convergence or better solutions, whereas \(\eta_{\alpha}\) and \(\eta_{\delta}\) are constants. In discrete optimization, \(a_{\text{max}}\) should be slightly larger than in comparison to continuous optimization. The main difference between GWO and IGWO is that they use dynamically varying functions. Also, other parts of IGWO are the same as GWO. Eventually, The IGWO uses the following equations for its updating strategy: 
\begin{equation}
    x_1 = x_{\alpha}^{i} - \left( 2 a_{\alpha}(i) \cdot r_1 - a_{\alpha}(i) \right) \cdot D_{\alpha},
\end{equation}
\begin{equation}
    x_2 = x_{\beta}^{i} - \left( 2 a_{\beta}(i) \cdot r_1 - a_{\beta}(i) \right) \cdot D_{\beta},
\end{equation}
\begin{equation}
  x_3 = x_{\delta}^{i} - \left( 2 a_{\delta}(i) \cdot r_1 - a_{\delta}(i) \right) \cdot D_{\delta},   
\end{equation}

\subsection*{Modified Gorilla Troops Optimizer (MGTO)}
The Modified Gorilla Troops Optimizer (MGTO)\cite{MOSTAFA2023110462} is an advanced version of this metaheuristic
algorithm of overcoming defects in the previous GTO\cite{abdollahzadeh2021artificial}, such as premature convergence and
low diversity of the population. To enhance its capability for exploration, MGTO has been
introduced with a technique known as Elite Opposition-Based Learning, which extends the
search space by considering the opposite solutions in order to cover a wider area and ensure
diversity during an early stage of the optimization process. Moreover, the Cauchy Inverse
Cumulative Distribution (CICD) operator contributes an adaptive mechanism of exploration
that will enable the algorithm to move through complex and high-dimensional landscapes
more effectively and find areas where improvements might potentially take place.
For the purpose of exploitation, MGTO leans on the TFO with the CICD operator in order to
fine-tune candidate solutions closer to optimal regions. These modifications further enhance
the strength of the algorithm for faster convergence without getting stuck in local optima.
Balancing effectively between the exploratory and exploitative phases of optimization,
MGTO tends to perform exceptionally well for a wide range of optimization scenarios
, from complex benchmark problems to real-world engineering applications.
Comparative analyses have shown the improvement in accuracy, stability, and convergence
efficiency that establishes MGTO as a powerful and reliable tool for challenging optimization
tasks.
\subsection*{Exploration Phase}

The exploration phase in the mGTO algorithm ensures the population diversifies by leveraging two operators: the Elite Opposition-Based Learning (EOBL) mechanism and the Cauchy Inverse Cumulative Distribution (CICD) operator.

 The EOBL mechanism generates an opposite position for each gorilla, improving the exploration capability of the algorithm. The new position is computed as:
\begin{equation}
x_k = y_k + z_k - x_k,
\end{equation}
or in a dynamic form:
\begin{equation}
x_{k,j} = F \cdot (d_{yj} + d_{zj}) - x_{k,j},
\end{equation}
where:
\begin{equation}
d_{yj} = \min(x_{k,j}), \quad d_{zj} = \max(x_{k,j}).
\end{equation}
If the newly computed position \(x_{k,j}\) exceeds the predefined bounds, it is adjusted using:
\begin{equation}
x_{k,j} = \text{rand}(y_j + z_j), \quad \text{if } x_{k,j} < y_j \text{ or } x_{k,j} > z_j.
\end{equation}
 To introduce controlled randomness, the CICD operator generates random values following the Cauchy distribution. This ensures a balance between small and large exploratory steps. The probability density function is defined as:
\begin{equation}
f(x; a, b) = \frac{1}{\pi} \cdot \frac{b}{(x-a)^2 + b^2},
\end{equation}
and its cumulative distribution function is:
\begin{equation}
F(x; a, b) = \frac{1}{\pi} \arctan\left(\frac{x-a}{b}\right) + \frac{1}{2}.
\end{equation}
The random variable \(x\) is then generated using:
\begin{equation}
x = a + b \cdot \tan\left(\pi \cdot \left(p - \frac{1}{2}\right)\right),
\end{equation}
where \(a\) and \(b\) are the position and scale parameters, respectively.

\subsection*{Exploitation Phase}

In the exploitation phase, gorillas refine their search by adjusting their positions relative to the silverback (the best solution found so far). Two scenarios guide their behavior, based on the parameter \(C\) compared to a threshold \(W\).

If (\(C \geq W\)): Gorillas adjust their positions toward the silverback by following:
\begin{equation}
X(t+1) = X(t) + L \cdot M \cdot (X(t) - X_{silverback}) \cdot (0.01 \cdot \tan(\pi \cdot (p - 0.5))),
\end{equation}
where \(M\) and \(L\) determine the convergence intensity.

If (\(C < W\)):In this case, the gorillas compete for dominance, and their movement is defined by:
\begin{equation}
X(i) = X_{silverback} - (X_{silverback} \cdot Q - X(t) \cdot Q) \cdot \tan(v \cdot \pi),
\end{equation}
where:
\begin{equation}
Q = 2 \cdot r_5 - 1, \quad A = \beta \cdot E,
\end{equation}
and:
\begin{equation}
E = 
\begin{cases}
N_1, & \text{if } \text{rand} \geq 0.5, \\
N_2, & \text{if } \text{rand} < 0.5.
\end{cases}
\end{equation}

The general position update rule for gorillas is as follows:
\begin{equation}
X(t+1) =
\begin{cases} 
(U_l - L_l) \cdot r_1 + L_l, & \text{if } \text{rand} < p, \\
(r_2 - C) \cdot X_r(t) + L \cdot H, & \text{if } \text{rand} \geq 0.5, \\
X(i) - L \cdot (L \cdot (X(t) - X_r(t)) + r_3 \cdot (X(t) - X_r(t))), & \text{otherwise}.
\end{cases}
\end{equation}

\begin{equation}
C = F \cdot \left(1 - \frac{It}{MaxIt}\right), \quad F = \cos(2 \cdot r_4) + 1, \quad L = C \cdot l,
\end{equation}
\begin{equation}
H = Z \cdot X(t),
\end{equation}
where \(r_1, r_2, r_3, r_4, r_5\) are random numbers within [0, 1], \(l\) is a random integer in \([-1, 1]\), and \(Z\) is a random value in \([-C, C]\).

\section*{Experiment}

The proposed model was implemented and trained on a system equipped with two NVIDIA RTX 4090, 256 GB of RAM, and running on a Linux Ubuntu operating system and the results have been compared with previous studies. For our research, we used datasets ISIC 2016 and ISIC 2017. More Descriptions of the datasets are available in the Data section. We used measures such as Accuracy, F-measure, Recall, and Precision to compare results. In the beginning, the metrics that have been used were described. Then, the evaluation results derived from these two databases are examined.

When implementing the method, the training samples from these databases were used to build the proposed model, and its performance was tested using a separate set of test samples. Key metrics like precision, accuracy, recall, and F-measure were applied to determine the method's effectiveness.

Once the test samples were classified by the proposed model (or other models for comparison), the predicted results were checked against the actual labels of the samples. Based on this comparison, each test sample fell into one of the following categories:

\begin{itemize}
    \item \textbf{True Positive (TP):} This means the model correctly identified a melanoma sample.
    \item \textbf{True Negative (TN):} This shows the model correctly identified a non-melanoma sample.
    \item \textbf{False Positive (FP):} This occurs when the model incorrectly classified a non-melanoma sample as melanoma.
    \item \textbf{False Negative (FN):} This happens when the model mistakenly classified a melanoma sample as non-melanoma.
\end{itemize}

This process helps us understand how well the model performs and where it might need improvement.

\subsection*{Data}
We used the ISIC 2016\cite{DBLP:journals/corr/GutmanCCHMMH16} and 2017\cite{DBLP:journals/corr/abs-1710-05006} datasets for our analysis. The ISIC 2016 and 2017 datasets are part of the International Skin Imaging Collaboration (ISIC) initiative, which aims to support the development of algorithms for the automated analysis of skin lesions, particularly for melanoma detection. The ISIC 2016 dataset includes around 900 high-quality dermoscopic images, mainly designed for lesion segmentation. Each image comes with a binary mask that marks the lesion's boundary, helping with tasks like identifying and segmenting skin lesions.\\
The ISIC 2017 dataset expands on this with over 2,000 images and is divided into three key tasks: lesion segmentation, dermoscopic feature extraction, and lesion classification. For segmentation, it also provides binary masks similar to the 2016 dataset. The feature extraction task focuses on identifying critical visual patterns like streaks, globules, and pigment networks, which are crucial for diagnosing melanoma. The classification part contains two categories: melanoma (374 images) and non-melanoma, which includes seborrheic keratosis (254 images) and benign nevi (1,372 images).

\subsection*{Accuracy Metrics}
In this work, the performance of the proposed skin cancer diagnostic model is evaluated by some key metrics defined below: 
\begin{itemize}
    \item \textbf{Accuracy}: 
    Accuracy measures the proportion of correctly classified cases (both positive and negative) to the total number of cases:
\begin{equation}
    \text{Accuracy} = \frac{\text{TP} + \text{TN}}{\text{TP} + \text{TN} + \text{FP} + \text{FN}}
\end{equation}
    \item \textbf{Precision}: 
    Precision reflects the reliability of positive predictions and is defined as:
    \begin{equation}
        \text{Precision} = \frac{\text{TP}}{\text{TP} + \text{FP}}
    \end{equation}

    \item \textbf{Recall (Sensitivity)}: 
    Recall measures the proportion of actual positive cases correctly identified:
    \begin{equation}
        \text{Recall} = \frac{\text{TP}}{\text{TP} + \text{FN}}
    \end{equation}

    \item \textbf{F-Measure}: 
    F-Measure is the harmonic mean of precision and recall, balancing their contributions:
    \begin{equation}
        \text{F-Measure} = 2 \times \frac{\text{Precision} \times \text{Recall}}{\text{Precision} + \text{Recall}}
    \end{equation}
    \item \textbf{Statistic}:
    The statistic is a numerical value derived from a statistical test that measures the difference between two groups or models. It serves as a tool to quantify the difference's magnitude and direction, providing insight into which group or model performs better. A positive statistic indicates that the first group or model outperforms the second, while a negative value suggests the opposite. The absolute value of the statistic reflects the strength of the observed difference—the larger the value, the more pronounced the disparity. However, while statistics help describe the extent of a difference, they do not determine whether the difference is statistically significant alone. For this reason, it is often paired with a p-value, which assesses the likelihood that the observed difference is due to chance. The statistic and p-value create a comprehensive framework for comparing groups or models, helping researchers determine whether observed differences are meaningful and reliable.
    \item \textbf{p-value}:
    The p-value is a statistical measure used to determine the significance of observed differences between two groups or models in a hypothesis test. It quantifies the probability that the observed results occurred purely by chance, assuming the null hypothesis is true. In our work, the null hypothesis typically states no significant difference in performance between the compared models. A smaller p-value (commonly less than 0.05) indicates that the observed difference is statistically significant and unlikely to have occurred by random chance, and the null hypothesis can be rejected. Conversely, a more significant p-value (more distinguished than 0.05) suggests that the difference could reasonably be due to random variation, and the null hypothesis cannot be rejected. The p-value complements other performance metrics by providing a rigorous, probabilistic framework for determining whether improvements in metrics like accuracy, precision, or recall are meaningful and not merely artifacts of noise or randomness. 
    
\end{itemize}

For the \textbf{p-value} and \textbf{Statsitc}, we use the t-test metric, which can be formulated as :
\begin{equation}
     t = \frac{\bar{X}_1 - \bar{X}_2}{\sqrt{\frac{s_1^2}{n_1} + \frac{s_2^2}{n_2}}} 
\end{equation}

In this formula, the variances of each group can be calculated by:
\begin{equation}
    s_1^2 = \frac{\sum_{i=1}^{n_1} (X_{1i} - \bar{X}_1)^2}{n_1 - 1}, \quad
s_2^2 = \frac{\sum_{i=1}^{n_2} (X_{2i} - \bar{X}_2)^2}{n_2 - 1}
\end{equation}

\section*{Results}

According to the results in table 1, \textbf{DenseNet} works better than other networks due to its dense connectivity. This characteristic of this network leads to more efficient feature reuse, by which both early-layer details and deeper-layer abstractions contribute to the final decision. Furthermore, DenseNet improves the Gradient flow \cite{he2016deep}; thus, the risk of vanishing gradient decreases. In addition, DenseNet reduces redundancy by generating fewer overlapping features, which leads to parameter efficiency and minimizes the risk of overfitting, making it more effective for datasets like ISIC, which may have limited samples. It can also preserve features across all layers that help the architecture to be robust for datasets with limited and imbalanced samples, like ISICs. However, DenseNet is less lightweight than the \textbf{MobileNet}, and if the computational resources are limited, the MobileNet can be a better option.

{\footnotesize
\begin{longtable}{llcccc}
\caption{Comparing The Efficiency of The Proposed Method With Different Methods} \label{tab:Test} \\ \hline
\textbf{Database} & \textbf{Method} & \textbf{Accuracy} & \textbf{F-measure} & \textbf{Recall} & \textbf{Precision} \\ \hline
\endfirsthead
\hline
\textbf{Database} & \textbf{Method} & \textbf{Accuracy} & \textbf{F-measure} & \textbf{Recall} & \textbf{Precision} \\ \hline
\endhead
\hline
\endfoot
\hline
\endlastfoot
\multirow{29}{*}{ISIC-2016} 
& Lai et al. \cite{Lai2023} & 97.0976 & 95.5817 & 97.1864 & 94.1755 \\ \cline{2-6}
& Nawaz et al.\cite{Nawaz2022} & 95.5145 & 93.2967 & 95.6974 & 91.3485 \\ \cline{2-6}
& InSiNet \cite{Reis2022} & 96.3061 & 94.4544 & 96.6930 & 92.6010 \\ \cline{2-6}
& Xception & 95.22 & 95.26 & 94.68 & 95.85 \\ \cline{2-6}
& Xception+Wavelet & 97.794 & 96.74 & 98.78 & 94.97 \\ \cline{2-6}
& Xception+Wavelet+IGWO & 97.773 & 97.74 & 98.56 & 96.94 \\ \cline{2-6}
& Xception+Wavelet+Fox & 96.92 & 97.85 & 98.35 & 97.36 \\ \cline{2-6}
& Xception+Wavelet+MGTO & 97.82 & 96.65 & 96.54 & 96.78 \\ \cline{2-6}
& Inception & 91.90 & 94.70 & 94.02 & 95.41 \\ \cline{2-6}
& Inception+Wavelet & 95.58 & 95.87 & 95.17 & 96.60 \\ \cline{2-6}
& Inception+Wavelet+IGWO & 96.65 & 96.63 & 97.17 & 96.10 \\ \cline{2-6}
& Inception+Wavelet+Fox & 97.75 & 97.74 & 97.70 & 97.80 \\ \cline{2-6}
& Inception+Wavelet+MGTO & 96.65 & 96.99 & 96.25 & 97.75 \\ \cline{2-6}
& DenseNet & 97.98 & 97.34 & 96.82 & 97.87 \\ \cline{2-6}
& DenseNet+Wavelet & 97.00 & 97.04 & 98.50 & 95.63 \\ \cline{2-6}
& DenseNet+Wavelet+IGWO & 98.01 & 97.76 & 98.80 & 96.66 \\ \cline{2-6}
& \textbf{DenseNet+Wavelet+Fox} & \textbf{98.11} & \textbf{98.39} & \textbf{99.01} & \textbf{97.78} \\ \cline{2-6}
& DenseNet+Wavelet+MGTO & 97.87 & 98.26 & 98.88 & 95.65 \\ \cline{2-6}
& MobileNet & 94.98 & 97.33 & 96.81 & 97.87 \\ \cline{2-6}
& MobileNet+Wavelet & 97.00 & 98.06 & 98.50 & 97.63 \\ \cline{2-6}
& MobileNet+Wavelet+IGWO & 97.01 & 98.27 & 98.90 & 97.66 \\ \cline{2-6}
& MobileNet+Wavelet+Fox & 98.11 & 97.88 & 99.01 & 96.89 \\ \cline{2-6}
& MobileNet+Wavelet+MGTO & 97.81 & 97.60 & 98.35 & 96.88 \\ \hline
\multirow{29}{*}{ISIC-2017} 
& Lai et al. \cite{Lai2023}& 95.1667 & 92.8606 & 96.35 & 90.2476 \\ \cline{2-6}
& Nawaz et al.\cite{Nawaz2022} & 93.1667 & 89.6725 & 91.8697 & 87.8916 \\ \cline{2-6}
& InSiNet \cite{Reis2022} & 94.6667 & 92.1440 & 95.7119 & 91.5320 \\ \cline{2-6}
& Xception & 95.85 & 95.87 & 95.17 & 96.60 \\ \cline{2-6}
& Xception+Wavelet & 96.35 & 96.35 & 96.50 & 96.21 \\ \cline{2-6}
& Xception+Wavelet+IGWO & 97.90 & 97.89 & 98.00 & 97.80 \\ \cline{2-6}
& Xception+Wavelet+Fox & 97.35 & 96.35 & 96.50 & 96.21 \\ \cline{2-6}
& Xception+Wavelet+MGTO & 97.89 & 97.89 & 98.00 & 97.80 \\ \cline{2-6}
& Inception & 96.20 & 96.25 & 94.85 & 97.70 \\ \cline{2-6}
& Inception+Wavelet & 95.90 & 96.01 & 99.00 & 93.21 \\ \cline{2-6}
& Inception+Wavelet+IGWO & 96.28 & 96.25 & 94.85 & 97.70 \\ \cline{2-6}
& Inception+Wavelet+Fox & 97.00 & 97.97 & 96.41 & 99.60 \\ \cline{2-6}
& \textbf{Inception+Wavelet+MGTO} & \textbf{97.95} & \textbf{97.97} & \textbf{96.41} & \textbf{99.60} \\ \cline{2-6}
& DenseNet & 96.85 & 96.85 & 97.20 & 96.52 \\ \cline{2-6}
& DenseNet+Wavelet & 96.85 & 96.85 & 97.20 & 96.52 \\ \cline{2-6}
& DenseNet+Wavelet+IGWO & 97.60 & 97.84 & 98.92 & 96.79 \\ \cline{2-6}
& DenseNet+Wavelet+Fox & 97.12 & 97.62 & 98.90 & 96.39 \\ \cline{2-6}
& DenseNet+Wavelet+MGTO & 97.85 & 96.55 & 97.10 & 96.11 \\ \cline{2-6}
& MobileNet & 96.80 & 97.95 & 99.01 & 96.81 \\ \cline{2-6}
& MobileNet+Wavelet & 97.01 & 97.72 & 99.09 & 96.40 \\ \cline{2-6}
& MobileNet+Wavelet+IGWO & 97.52 & 97.56 & 98.30 & 96.84 \\ \cline{2-6}
& MobileNet+Wavelet+Fox & 97.81 & 97.61 & 98.36 & 96.89 \\ \cline{2-6}
& MobileNet+Wavelet+MGTO & 97.81 & 97.60 & 98.35 & 96.88 \\ \hline

\end{longtable}
}

\begin{figure}[ht]
    \centering
    \begin{minipage}[t]{0.45\textwidth}
        \includegraphics[width=6cm, height=4.5cm]{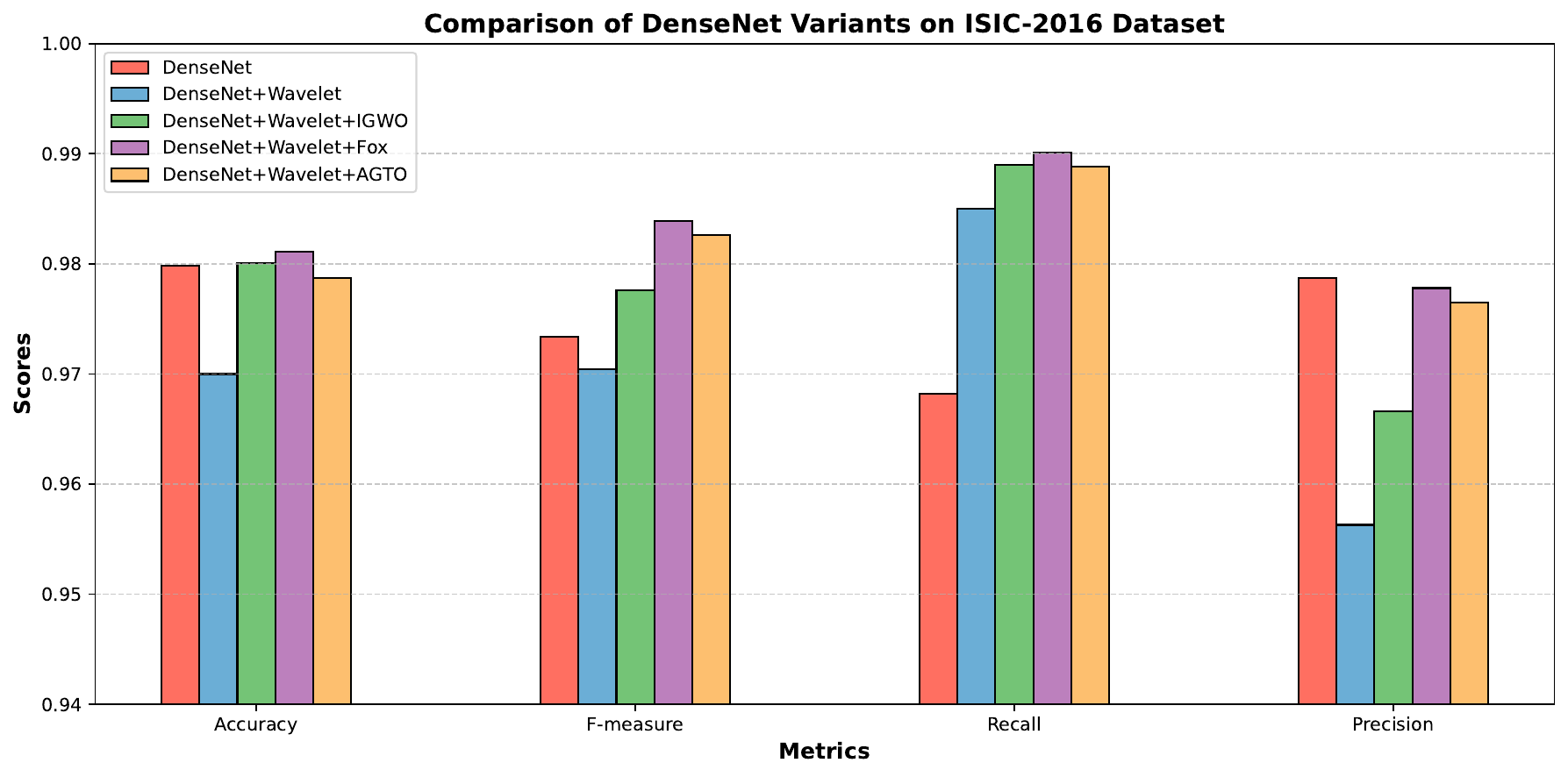}\\
        \centering DenseNet
    \end{minipage}
    \hfill
    \begin{minipage}[t]{0.45\textwidth}
        \includegraphics[width=6cm, height=4.5cm]{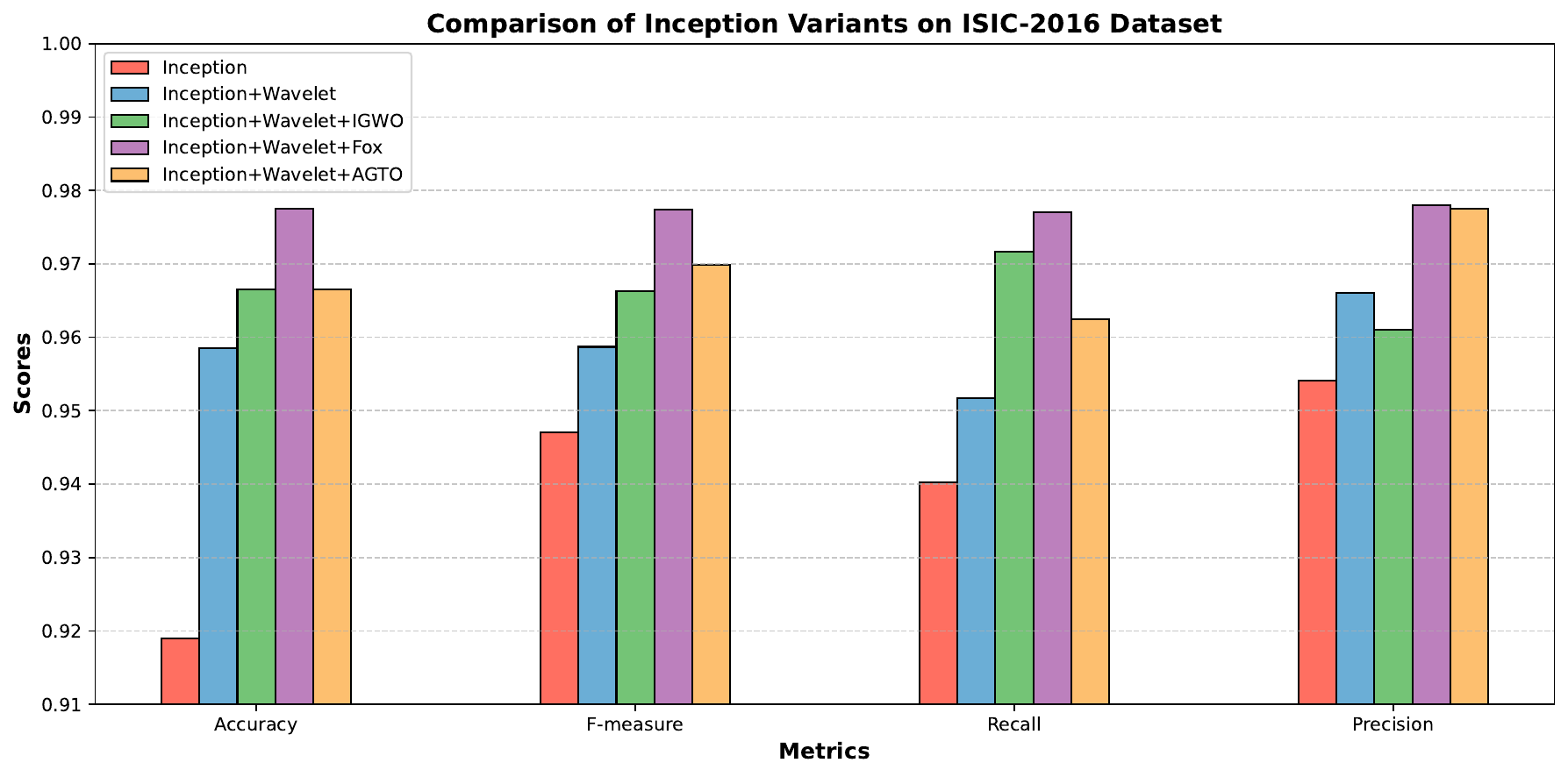}\\
        \centering Inception
    \end{minipage}

    \vspace{1em}

    \begin{minipage}[t]{0.45\textwidth}
        \includegraphics[width=6cm, height=4.5cm]{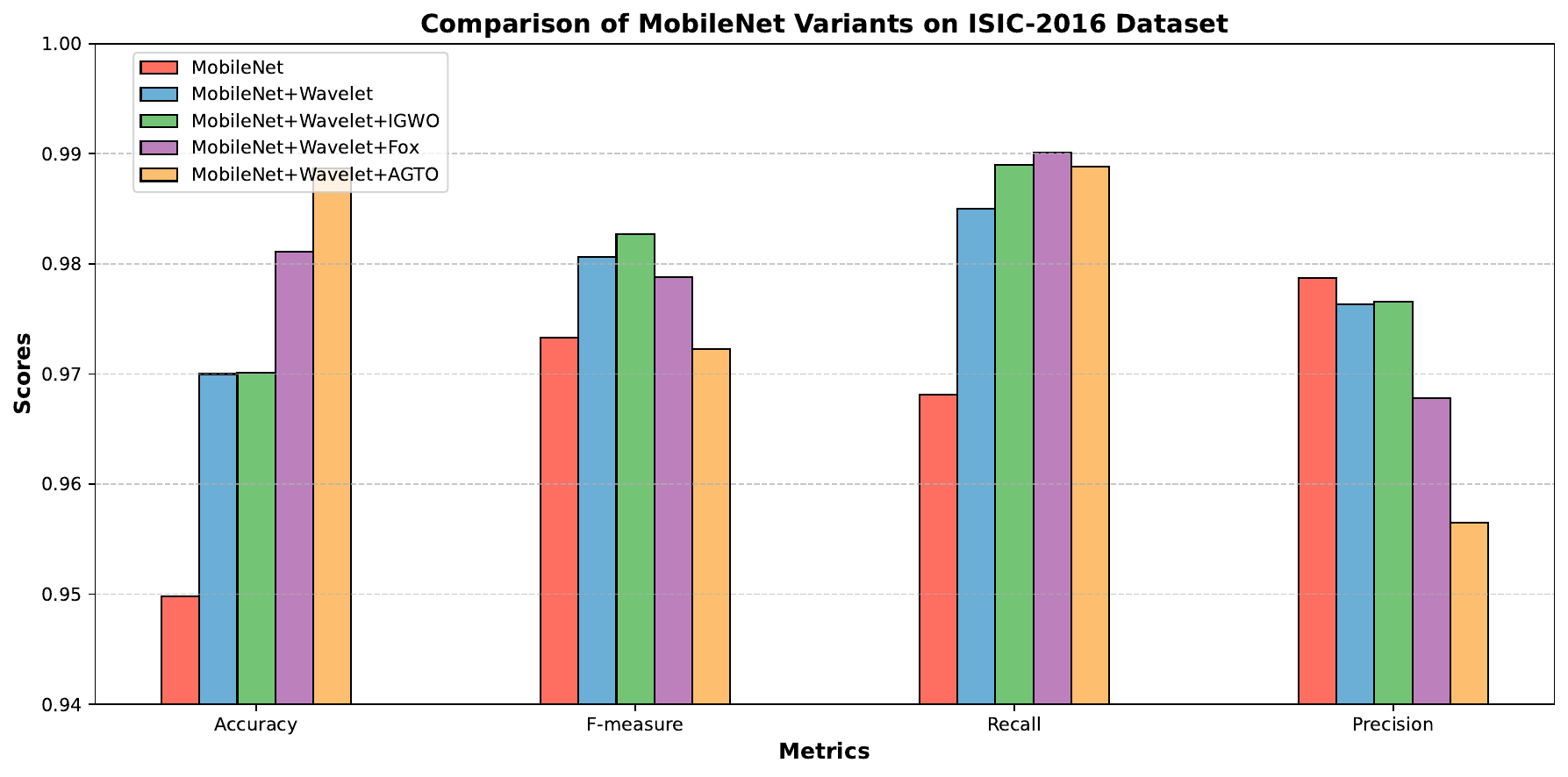}\\
        \centering MobileNet
    \end{minipage}
    \hfill
    \begin{minipage}[t]{0.45\textwidth}
        \includegraphics[width=6cm, height=4.5cm]{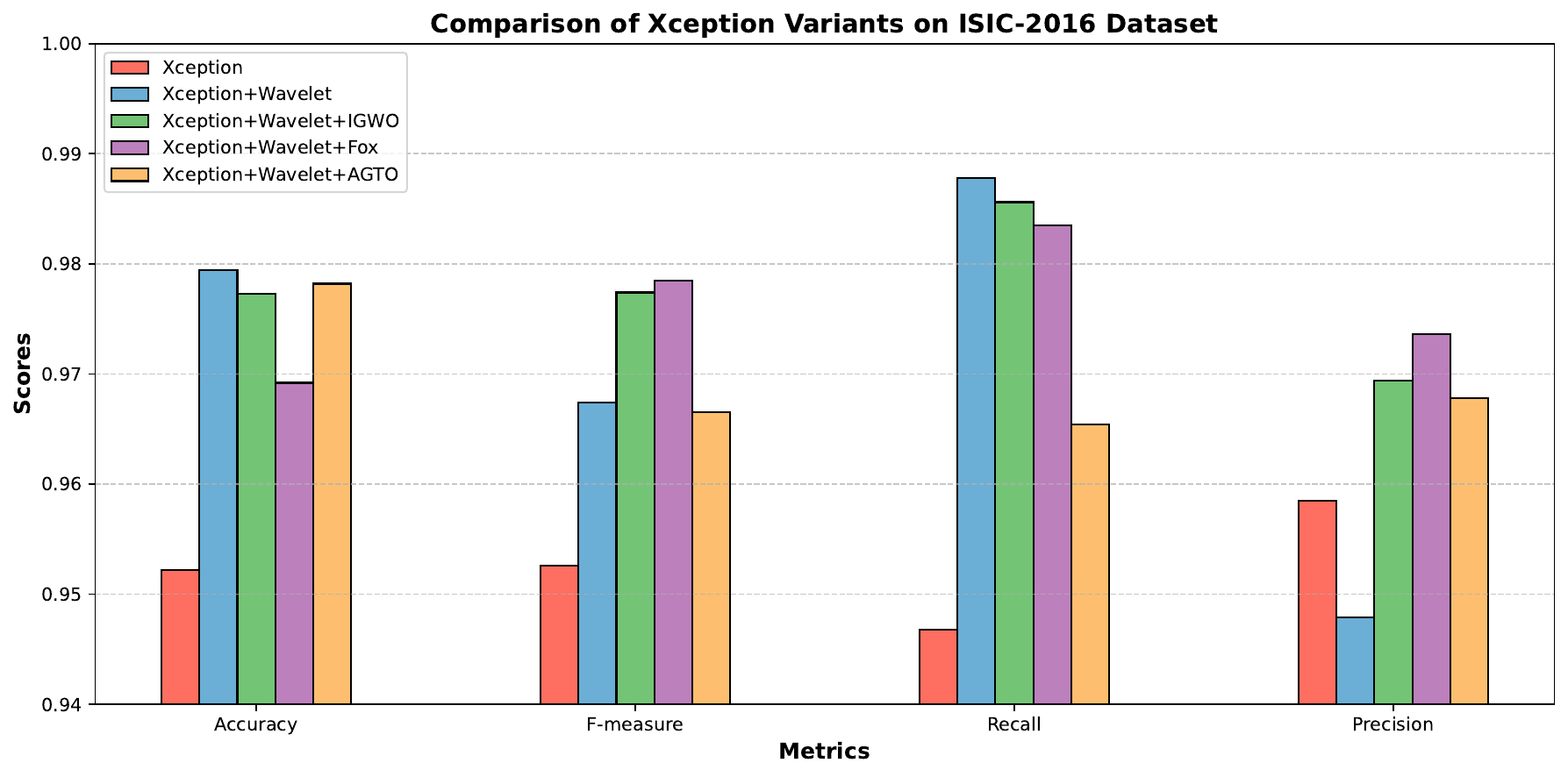}\\
        \centering Xception
    \end{minipage}

    \caption{Bar plot for accuracy of different models on the classification of ISIC-2016 dataset.}
    \label{fig:architectures} 
\end{figure}

\begin{figure}[ht]
    \centering
    \begin{minipage}[t]{0.45\textwidth}
        \includegraphics[width=6cm, height=4.5cm]{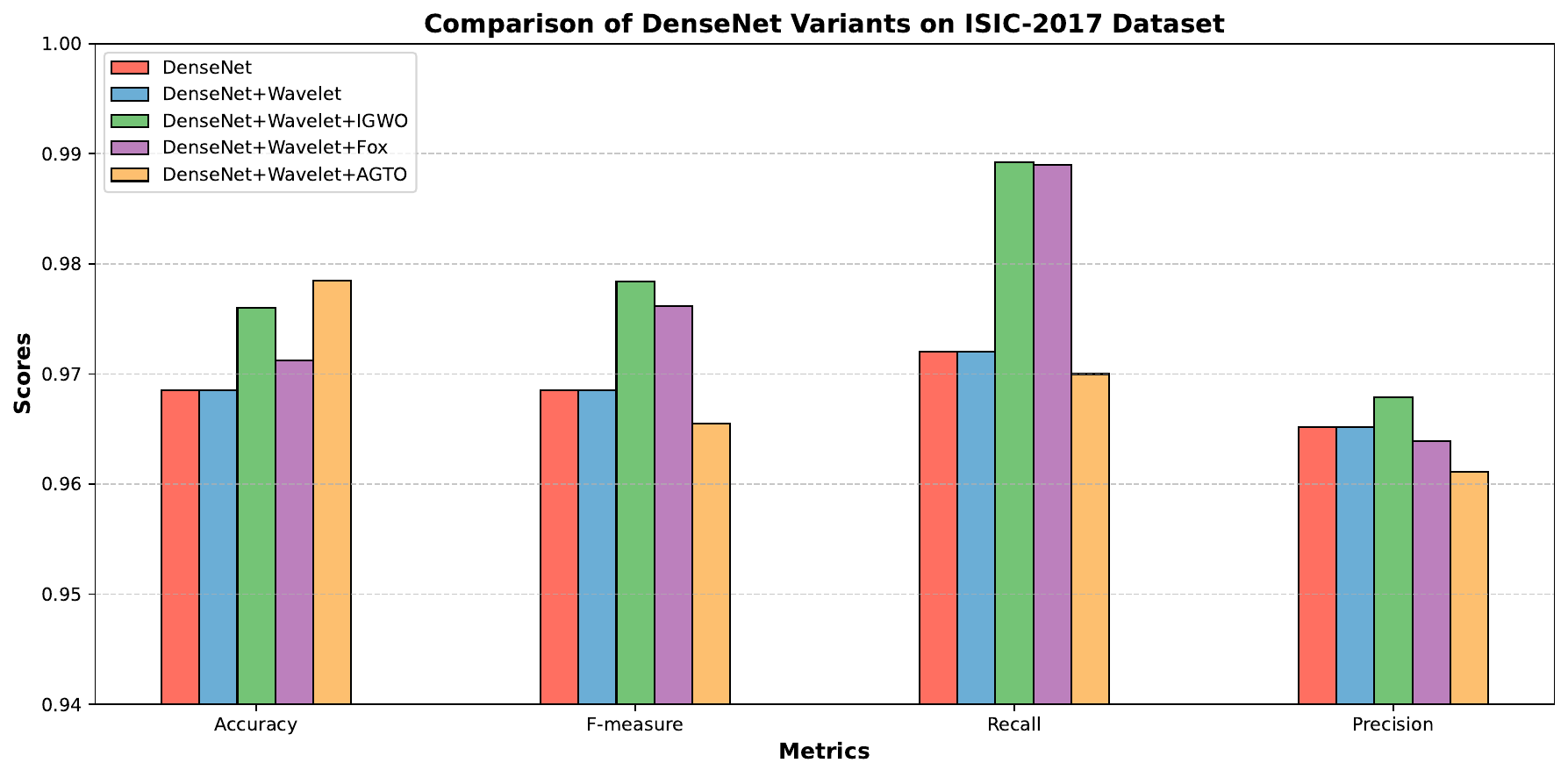}\\
        \centering DenseNet
    \end{minipage}
    \hfill
    \begin{minipage}[t]{0.45\textwidth}
        \includegraphics[width=6cm, height=4.5cm]{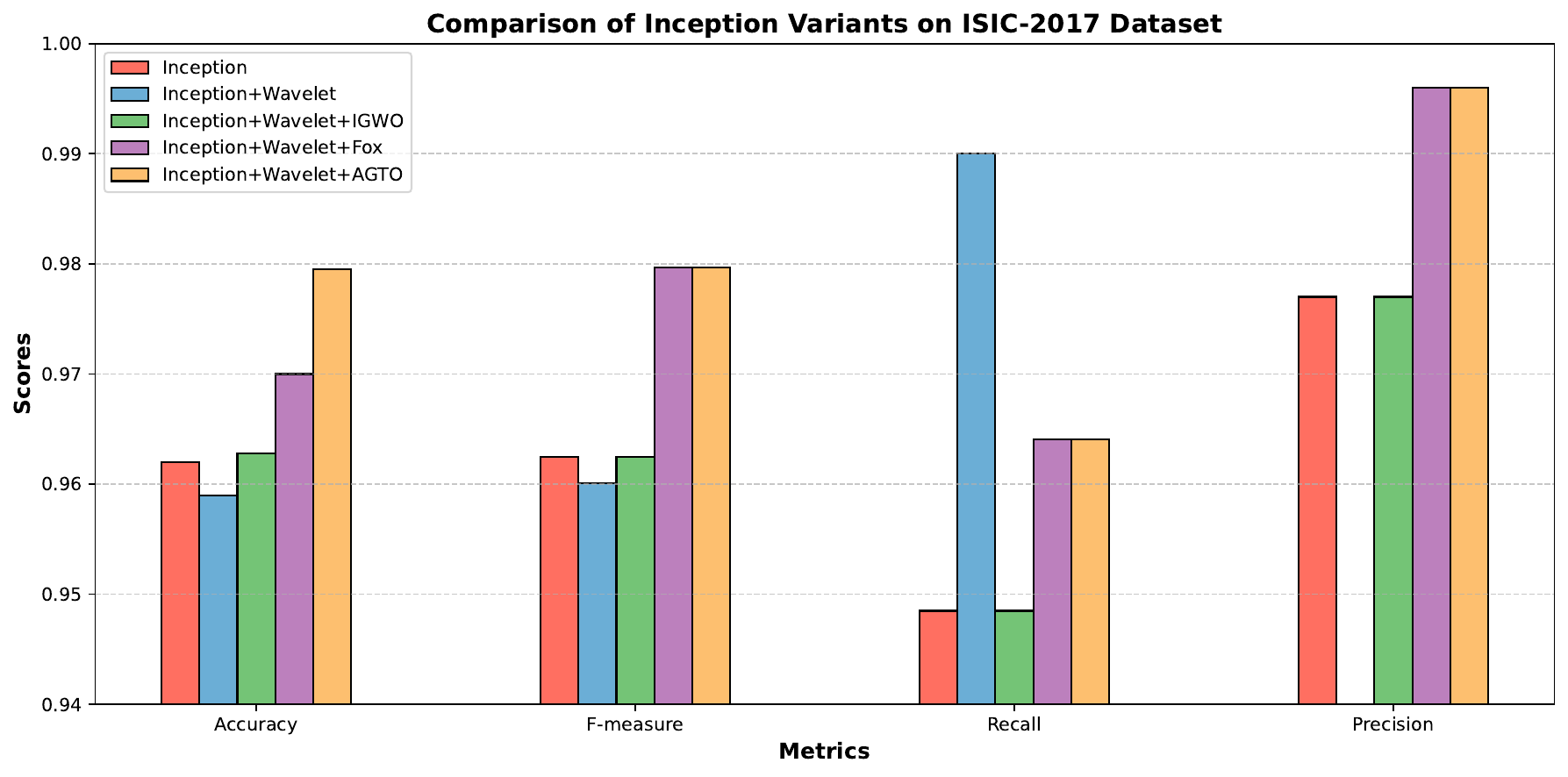}\\
        \centering Inception
    \end{minipage}

    \vspace{1em}

    \begin{minipage}[t]{0.45\textwidth}
        \includegraphics[width=6cm, height=4.5cm]{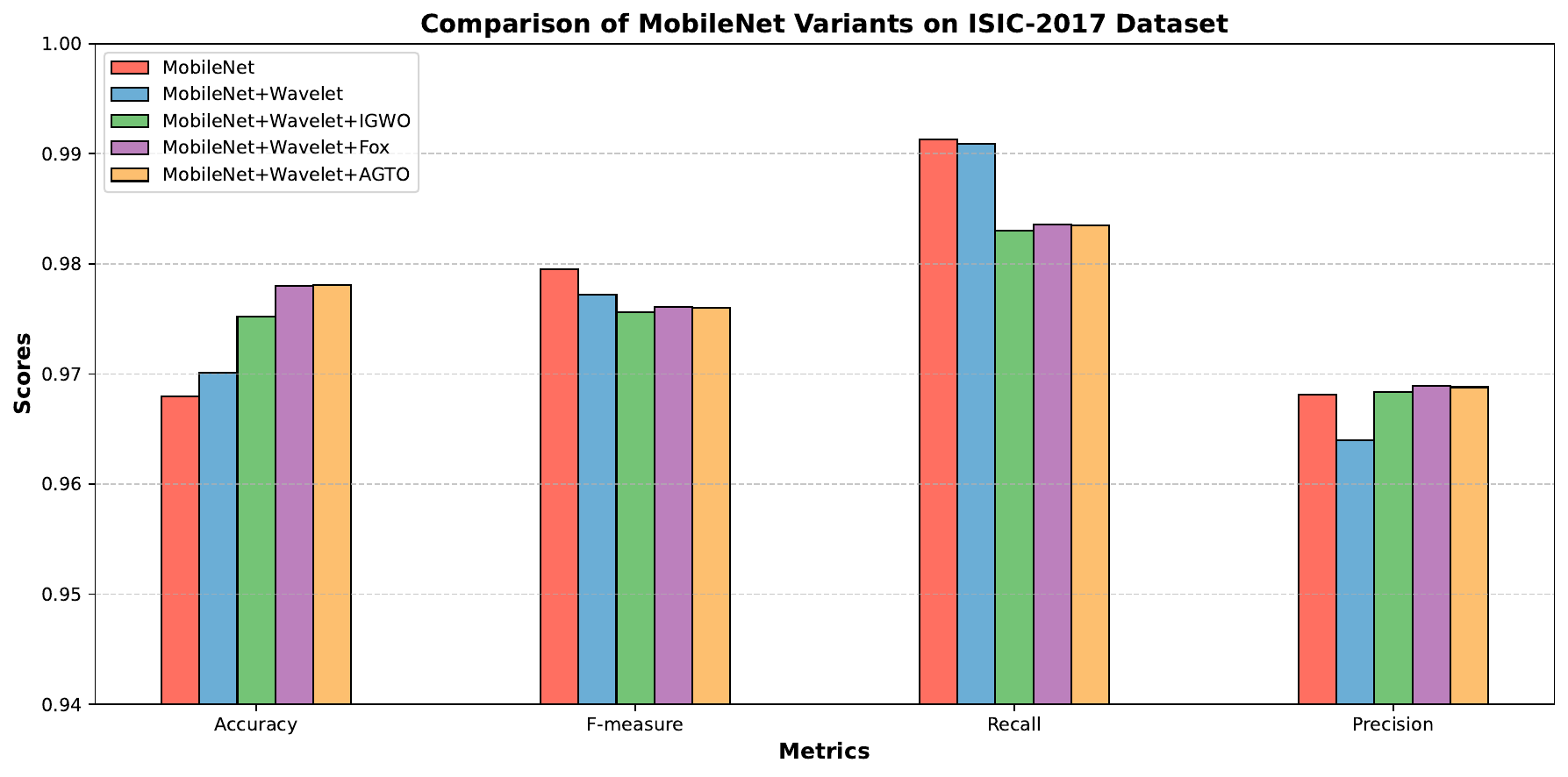}\\
        \centering MobileNet
    \end{minipage}
    \hfill
    \begin{minipage}[t]{0.45\textwidth}
        \includegraphics[width=6cm, height=4.5cm]{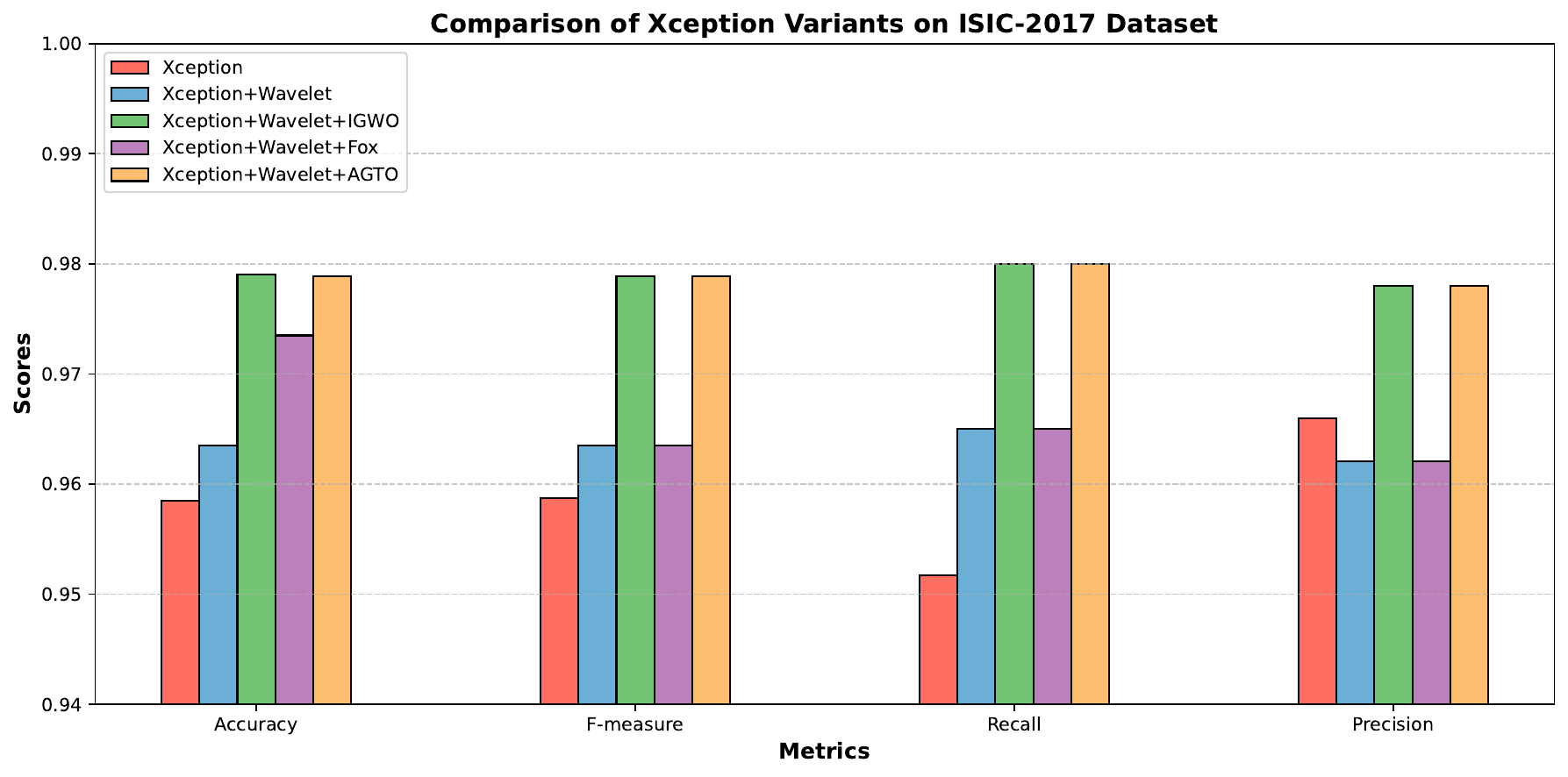}\\
        \centering Xception
    \end{minipage}

    \caption{Bar plot for accuracy of different models on the classification of ISIC-2017 dataset.}
    \label{fig:architectures} 
\end{figure}

\clearpage

According to the tables \ref{tab:Fold16}, \ref{tab:Fold17}, the k-fold cross-validation has been tested on the train section of the dataset ISIC-2016, which means that 15\% of the train data is assumed as a validation section and the other 65\% is used as a train data. As the table shows, the architectures are robust to different data from various parts of the dataset, demonstrating how great the architecture is. Furthermore, the accuracy of each model is well enough to show that the architecture parameters are determined quite well. Consequently, there is no sign of overfitting and underfitting in these tables. 

\begin{table}[ht]
\centering
\caption{Model Performance Across 5 Folds based on ISIC-2016} 
\begin{tabular}{lccccc}
\toprule
\textbf{Model} & \textbf{Fold1} & \textbf{Fold2} & \textbf{Fold3} & \textbf{Fold4} & \textbf{Fold5} \\
\midrule
Xception & 0.9511 & 0.9421 & 0.9233 & 0.9540 & 0.9544 \\
Xception+Wavelet & 0.9712 & 0.9695 & 0.9655 & 0.9705 & 0.9712 \\
Xception+Wavelet+IGWO & 0.9715 & 0.9754 & 0.9665 & 0.9623 & 0.9627 \\
Xception+Wavelet+Fox & 0.9802 & 0.9700 & 0.9569 & 0.9613 & 0.9725 \\
Xception+Wavelet+MGTO & 0.9811 & 0.9699 & 0.9666 & 0.9643 & 0.9715 \\
Inception & 0.9601 & 0.9632 & 0.9651 & 0.9501 & 0.9391 \\
Inception+Wavelet & 0.9630 & 0.9630 & 0.9614 & 0.9743 & 0.9698 \\
Inception+Wavelet+IGWO & 0.9790 & 0.9712 & 0.9743 & 0.9719 & 0.9716 \\
Inception+Wavelet+Fox & 0.9721 & 0.9703 & 0.9703 & 0.9718 & 0.9756 \\
Inception+Wavelet+MGTO & 0.9787 & 0.9776 & 0.9776 & 0.9719 & 0.9756 \\
DenseNet & 0.9701 & 0.9700 & 0.9702 & 0.9721 & 0.9753 \\
DenseNet+Wavelet & 0.9756 & 0.9690 & 0.9678 & 0.9612 & 0.9856 \\
DenseNet+Wavelet+IGWO & 0.9782 & 0.9782 & 0.9714 & 0.9781 & 0.9851 \\
DenseNet+Wavelet+Fox & 0.9743 & 0.9723 & 0.9723 & 0.9683 & 0.9813 \\
DenseNet+Wavelet+MGTO & 0.9744 & 0.9712 & 0.9710 & 0.9676 & 0.9823 \\
MobileNet & 0.9709 & 0.9800 & 0.9713 & 0.9687 & 0.9780 \\
MobileNet+Wavelet & 0.9809 & 0.9712 & 0.9710 & 0.9802 & 0.9702 \\
MobileNet+Wavelet+IGWO & 0.9831 & 0.9804 & 0.9810 & 0.9810 & 0.9813 \\
MobileNet+Wavelet+Fox & 0.9876 & 0.9612 & 0.9700 & 0.9822 & 0.9743 \\
MobileNet+Wavelet+MGTO & 0.9811 & 0.9604 & 0.9805 & 0.9799 & 0.9823 \\
\bottomrule
\end{tabular}
\label{tab:Fold16}
\end{table}

\begin{figure}[h]
    \centering
    \begin{minipage}[t]{0.45\textwidth}
        \includegraphics[width=8cm, height=4cm]{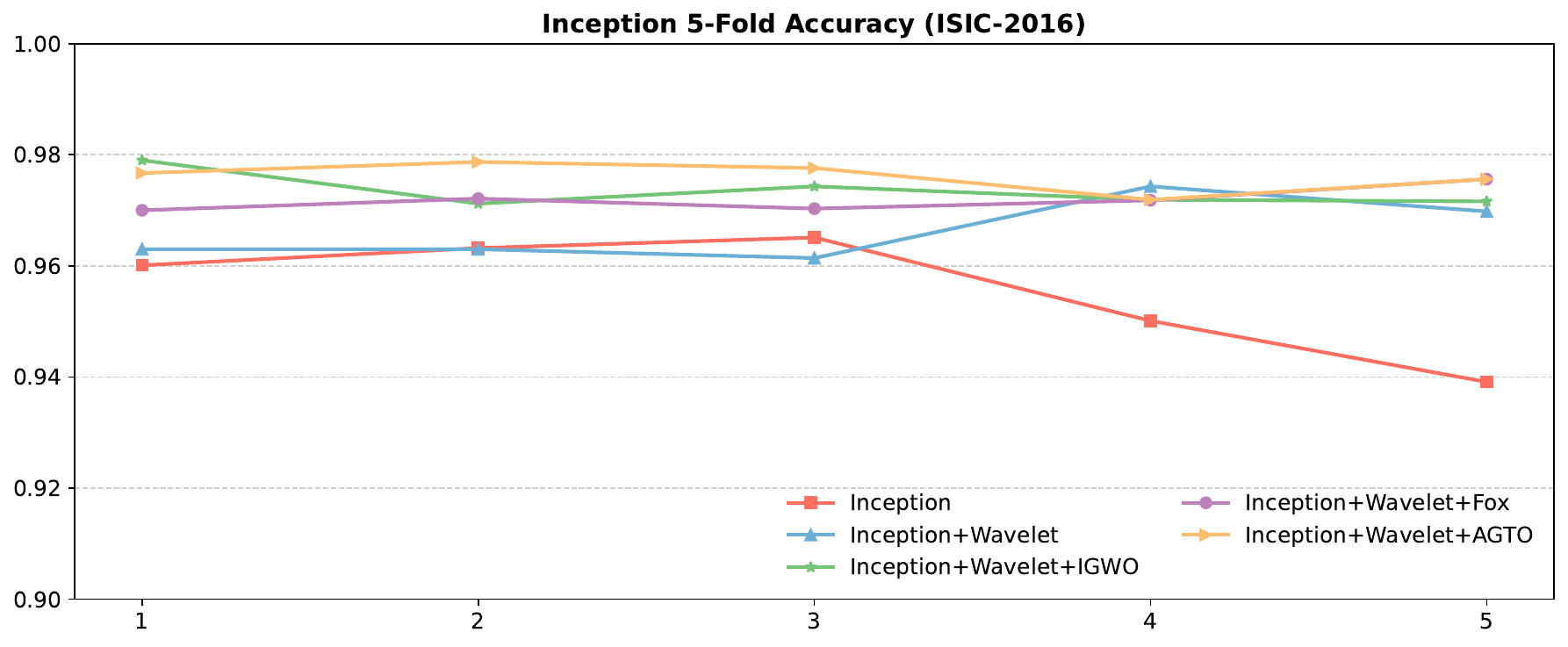}\\
        \centering DenseNet
    \end{minipage}
    \hfill
    \begin{minipage}[t]{0.45\textwidth}
        \includegraphics[width=8cm, height=4cm]{InceptionFold2016.pdf}\\
        \centering Inception
    \end{minipage}

    \vspace{1em}

    \begin{minipage}[t]{0.45\textwidth}
        \includegraphics[width = 8cm, height=4cm]{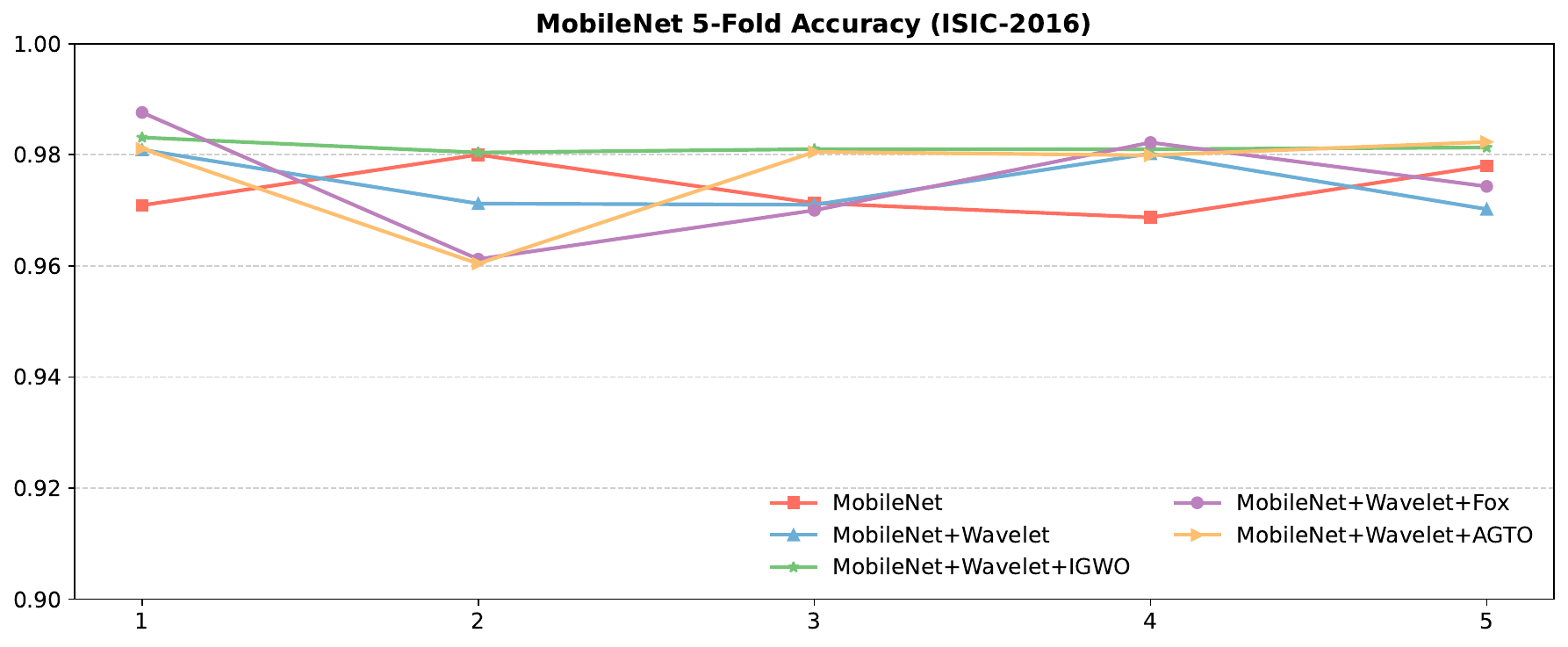}\\
        \centering MobileNet
    \end{minipage}
    \hfill
    \begin{minipage}[t]{0.45\textwidth}
        \includegraphics[width=8cm, height=4cm]{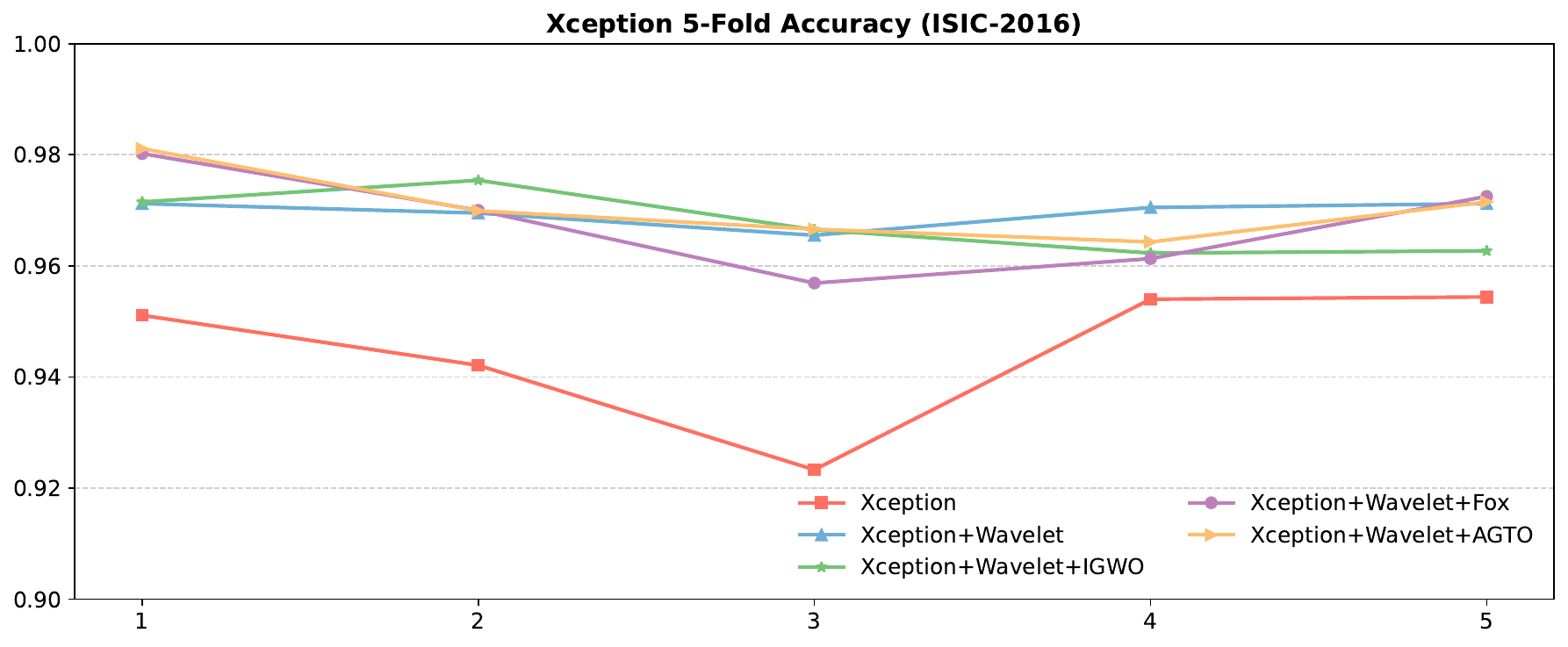}\\
        \centering Xception
    \end{minipage}

    \caption{Diagram for Model Performance Across 5 Folds on ISIC-2016 Dataset.}
    \label{fig:photos16} 
\end{figure}

According to table \ref{tab:xception_statistic}, adding a wavelet after the network enhances the model's performance a lot, which can be seen by the small amount of p-value(less than 0.05). However, adding an optimizer after the wavelet does not influence the model complexity greatly because the p-values are very considerable, which shows that the statistical difference is low if any optimizer is utilized after the wavelet transform. Conversely, according to table \ref{tab:inception_statistic}, adding a wavelet transformer alone after the Inception model does not affect the performance as much as adding an optimizer after the wavelet transformer does. This difference can also be observed in the table \ref{tab:Test}. Similar to table \ref{tab:inception_statistic}, tables \ref{tab:densenet_statistic},  \ref{tab:mobilenet_statistic} shows that adding a wavelet transformer without an optimizer does not enhance the performance as well as adding an optimizer, particularly IGWO.

\begin{table}[h]
\clearpage
\centering
\caption{Xception statistic and pvalue (ISIC-2016)}
\resizebox{\textwidth}{!}{%
\begin{tabular}{lccccc}
\toprule
\textbf{Model} & \textbf{Xception} & \textbf{Xception+Wavelet} & \textbf{Xception+Wavelet+IGWO} & \textbf{Xception+Wavelet+Fox} & \textbf{Xception+Wavelet+MGTO} \\
\midrule
Xception & 0.0, 1.0 & -4.1328, 0.0032 & -3.5555, 0.0074 & -3.2377, 0.01191 & -3.9350, 0.0043 \\
Xception+Wavelet & 4.1328, 0.0032 & 0.0, 1.0 & 0.6890, 0.5102 & 0.3282, 0.7511 & -0.3569, 0.7303 \\
Xception+Wavelet+IGWO & 3.5555, 0.0074 & -0.6890, 0.5102 & 0.0, 1.0 & -0.10310, 0.9204 & -0.7791, 0.4583 \\
Xception+Wavelet+Fox & 3.2377, 0.01191 & -0.3282, 0.7511 & 0.1031, 0.9204 & 0.0, 1.0 & -0.4959, 0.6332 \\
Xception+Wavelet+MGTO & 3.9350, 0.0043 & 0.3569, 0.7303 & 0.7791, 0.4583 & 0.4959, 0.6332 & 0.0, 1.0 \\
\bottomrule
\end{tabular}%
}
\label{tab:xception_statistic}
\end{table}
\begin{table}[h]
\centering
\caption{Inception statistic and pvalue (ISIC-2016)}
\resizebox{\textwidth}{!}{%
\begin{tabular}{lccccc}
\toprule
\textbf{Model} & \textbf{Inception} & \textbf{Inception+Wavelet} & \textbf{Inception+Wavelet+IGWO} & \textbf{Inception+Wavelet+Fox} & \textbf{Inception+Wavelet+MGTO} \\
\midrule
Inception & 0.0, 1.0 & -1.9801, 0.0830 & -3.5697, 0.0072 & -3.3192, 0.0105 & -4.1243, 0.0033 \\
Inception+Wavelet & 1.9801, 0.0830 & 0.0, 1.0 & -2.5466, 0.0343 & -2.1246, 0.0663 & -3.5868, 0.0071 \\
Inception+Wavelet+IGWO & 3.5697, 0.0072 & 2.5466, 0.0343 & 0.0, 1.0 & 0.9300, 0.3795 & -1.3405, 0.2168 \\
Inception+Wavelet+Fox & 3.3192, 0.0105 & 2.1246, 0.0663 & -0.9300, 0.3795 & 0.0, 1.0 & -2.6962, 0.0272 \\
Inception+Wavelet+MGTO & 4.1243, 0.0033 & 3.5868, 0.0071 & 1.3405, 0.2168 & 2.6962, 0.0272 & 0.0, 1.0 \\
\bottomrule
\end{tabular}
}
\label{tab:inception_statistic}
\end{table}
\begin{table}[h]
\centering
\caption{DenseNet statistic and pvalue (ISIC-2016)}
\resizebox{\textwidth}{!}{%
\begin{tabular}{lccccc}
\toprule
\textbf{Model} & \textbf{DenseNet} & \textbf{DenseNet+Wavelet} & \textbf{DenseNet+Wavelet+IGWO} & \textbf{DenseNet+Wavelet+Fox} & \textbf{DenseNet+Wavelet+MGTO} \\
\midrule
DenseNet & 0.0, 1.0 & -0.0705, 0.9454 & -2.3359, 0.0477 & -1.0857, 0.3092 & -0.6534, 0.5318 \\
DenseNet+Wavelet & 0.0705, 0.9454 & 0.0, 1.0 & -1.1801, 0.2718 & -0.4831, 0.6419 & -0.3026, 0.7698 \\
DenseNet+Wavelet+IGWO & 2.3359, 0.0477 & 1.1801, 0.2718 & 0.0, 1.0 & 1.0702, 0.3157 & 1.2137, 0.2594 \\
DenseNet+Wavelet+Fox & 1.0857, 0.3092 & 0.4831, 0.6419 & -1.0702, 0.3157 & 0.0, 1.0 & 0.2389, 0.8171 \\
DenseNet+Wavelet+MGTO & 0.6534, 0.5318 & 0.3026, 0.7698 & -1.2137, 0.2594 & -0.2389, 0.8171 & 0.0, 1.0 \\
\bottomrule
\end{tabular}
}
\label{tab:densenet_statistic}
\end{table}

\begin{table}[h!]
\centering
\caption{MobileNet statistic and pvalue (ISIC-2016)}
\resizebox{\textwidth}{!}{%
\begin{tabular}{lccccc}
\toprule
\textbf{Model} & \textbf{MobileNet} & \textbf{MobileNet+Wavelet} & \textbf{MobileNet+Wavelet+IGWO} & \textbf{MobileNet+Wavelet+Fox} & \textbf{MobileNet+Wavelet+MGTO} \\
\midrule
MobileNet & 0.0, 1.0 & -0.2828, 0.7844 & -3.3737, 0.0097 & -0.2502, 0.8087 & -0.6540, 0.5314 \\
MobileNet+Wavelet & 0.2828, 0.7844 & 0.0, 1.0 & -2.7292, 0.0258 & -0.0691, 0.9465 & -0.4482, 0.6658 \\
MobileNet+Wavelet+IGWO & 3.3737, 0.0097 & 2.7292, 0.0258 & 0.0, 1.0 & 1.3575, 0.2116 & 1.0879, 0.3082 \\
MobileNet+Wavelet+Fox & 0.2502, 0.8087 & 0.0691, 0.9465 & -1.3575, 0.2116 & 0.0, 1.0 & -0.2873, 0.7811 \\
MobileNet+Wavelet+MGTO & 0.6540, 0.5314 & 0.4482, 0.6658 & -1.0879, 0.3082 & 0.2873, 0.7811 & 0.0, 1.0 \\
\bottomrule
\end{tabular}
}
\label{tab:mobilenet_statistic}
\end{table}

\begin{table}[h!]
\centering
\caption{Model Performance Across 5 Folds (ISIC-2017)}
\begin{tabular}{lccccc}
\toprule
\textbf{Model} & \textbf{Fold1} & \textbf{Fold2} & \textbf{Fold3} & \textbf{Fold4} & \textbf{Fold5} \\
\midrule
Xception & 0.9515 & 0.9190 & 0.9265 & 0.9665 & 0.9080 \\
Xception+Wavelet & 0.9510 & 0.9175 & 0.9585 & 0.9630 & 0.9625 \\
Xception+Wavelet+IGWO & 0.9665 & 0.9612 & 0.9625 & 0.9775 & 0.9625 \\
Xception+Wavelet+Fox & 0.9762 & 0.9676 & 0.9622 & 0.9776 & 0.9623 \\
Xception+Wavelet+MGTO & 0.9776 & 0.9653 & 0.9665 & 0.9767 & 0.9646 \\
Inception & 0.9311 & 0.9361 & 0.9791 & 0.9125 & 0.9515 \\
Inception+Wavelet & 0.9355 & 0.9460 & 0.9700 & 0.9105 & 0.9510 \\
Inception+Wavelet+IGWO & 0.9510 & 0.9690 & 0.9680 & 0.9590 & 0.9505 \\
Inception+Wavelet+Fox & 0.9725 & 0.9795 & 0.9700 & 0.9680 & 0.9690 \\
Inception+Wavelet+MGTO & 0.9771 & 0.9715 & 0.9700 & 0.9701 & 0.9791 \\
DenseNet & 0.9590 & 0.9505 & 0.9523 & 0.9414 & 0.9570 \\
DenseNet+Wavelet & 0.9653 & 0.9555 & 0.9543 & 0.9423 & 0.9523 \\
DenseNet+Wavelet+IGWO & 0.9691 & 0.9523 & 0.9512 & 0.9643 & 0.9575 \\
DenseNet+Wavelet+Fox & 0.9545 & 0.9674 & 0.9620 & 0.9605 & 0.9523 \\
DenseNet+Wavelet+MGTO & 0.9700 & 0.9650 & 0.9690 & 0.9689 & 0.9723 \\
MobileNet & 0.9300 & 0.9235 & 0.9610 & 0.9435 & 0.9585 \\
MobileNet+Wavelet & 0.9412 & 0.9333 & 0.9620 & 0.9423 & 0.9553 \\
MobileNet+Wavelet+IGWO & 0.9498 & 0.9498 & 0.9672 & 0.9443 & 0.9524 \\
MobileNet+Wavelet+Fox & 0.9585 & 0.9590 & 0.9590 & 0.9620 & 0.9680 \\
MobileNet+Wavelet+MGTO & 0.9680 & 0.9680 & 0.9675 & 0.9788 & 0.9755 \\
\bottomrule
\end{tabular}
\label{tab:Fold17}
\end{table}

\begin{figure}[ht]
    \centering
    \begin{minipage}[t]{0.45\textwidth}
        \includegraphics[width=8cm, height=4cm]{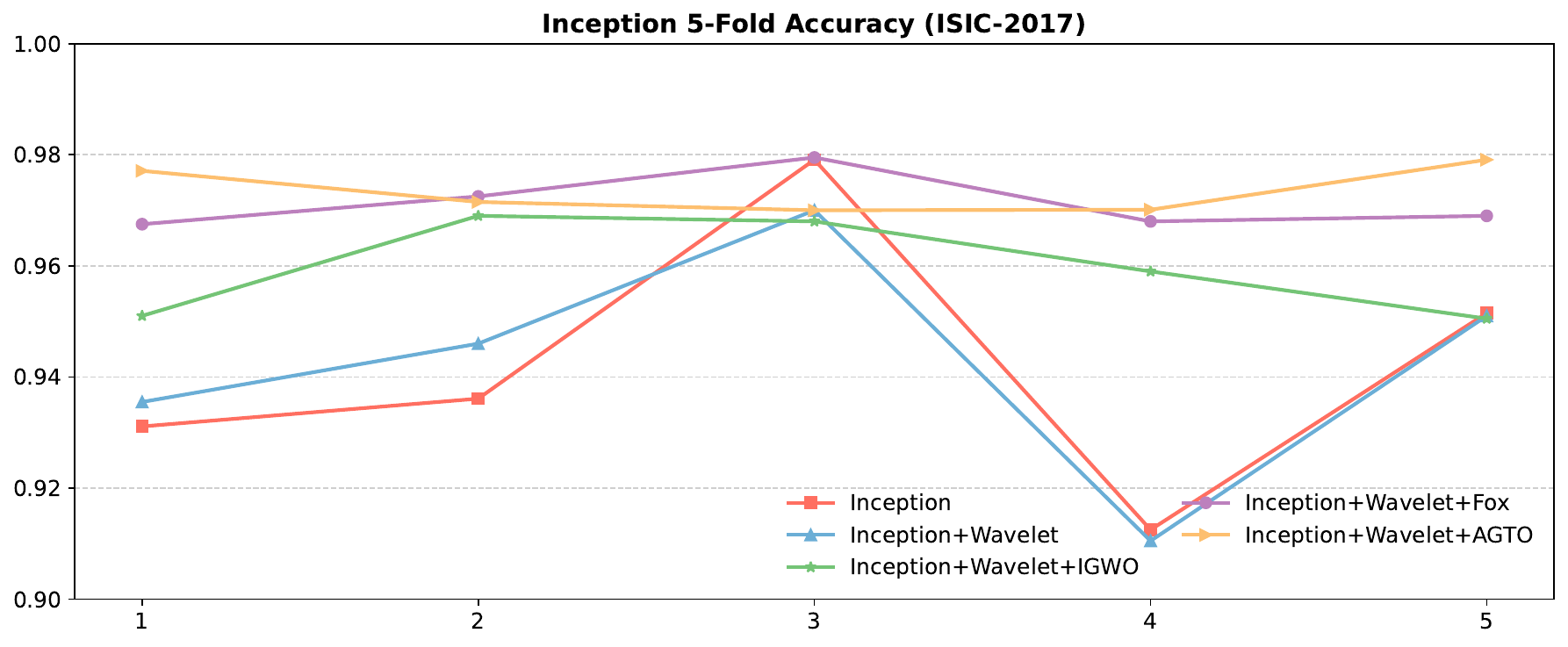}\\
        \centering DenseNet
    \end{minipage}
    \hfill
    \begin{minipage}[t]{0.45\textwidth}
        \includegraphics[width=8cm, height=4cm]{InceptionFold2017.pdf}\\
        \centering Inception
    \end{minipage}

    \vspace{1em}

    \begin{minipage}[t]{0.45\textwidth}
        \includegraphics[width = 8cm, height=4cm]{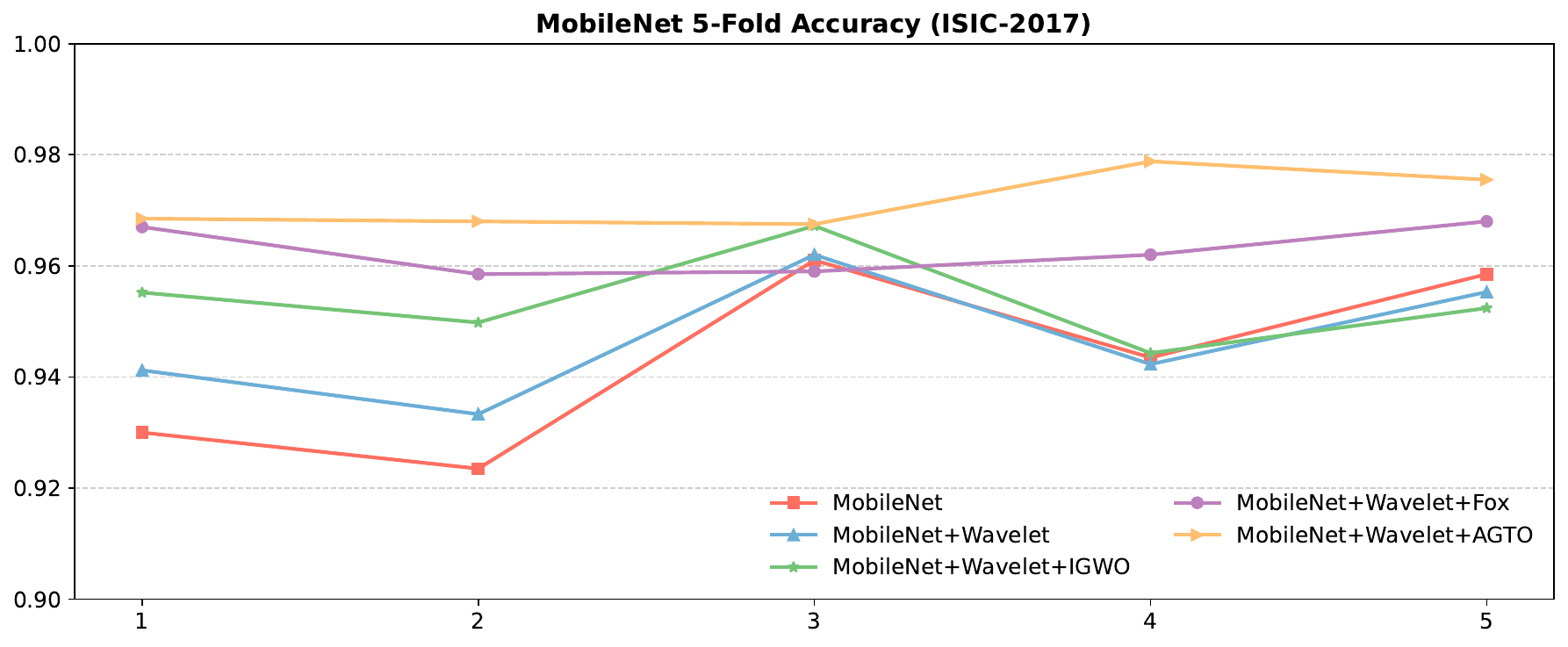}\\
        \centering MobileNet
    \end{minipage}
    \hfill
    \begin{minipage}[t]{0.45\textwidth}
        \includegraphics[width=8cm, height=4cm]{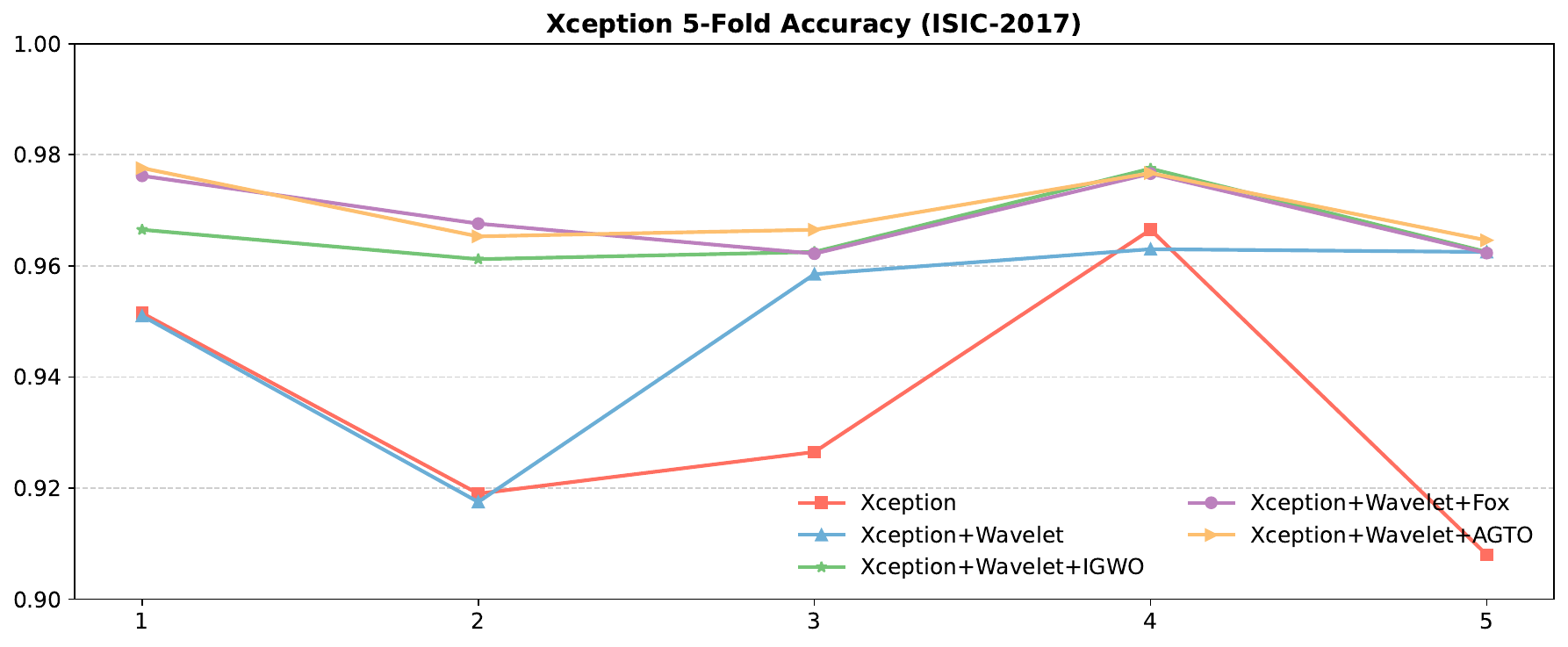}\\
        \centering Xception
    \end{minipage}

    \caption{Diagram for Model Performance Across 5 Folds on ISIC-2017 Dataset.}
    \label{fig:architectures} 
\end{figure}

\clearpage

 Like the previous tables \ref{tab:xception_statistic} to \ref{tab:mobilenet_statistic}, according to table \ref{tab:xception_statistic17}, connecting a wavelet transform after the base model enhances the model significantly due to the small amount of p-value (less than 0.05), however, adding an optimizer does not have a considerable impact on the model+wavelet. Conversely, it is clear in the tables \ref{tab:inception_statistic17},\ref{tab:densenet_statistic17}, \ref{tab:mobilenet_statistic17} that adding an optimizer, particularly IGWO, leads to better performance concerning the model+wavelet. 

\begin{table}[h!]
\centering
\caption{Xception statistic and pvalue (ISIC-2017)}
\resizebox{\textwidth}{!}{%
\begin{tabular}{lccccc}
\toprule
\textbf{Model} & \textbf{Xception} & \textbf{Xception+Wavelet} & \textbf{Xception+Wavelet+IGWO} & \textbf{Xception+Wavelet+Fox} & \textbf{Xception+Wavelet+MGTO} \\
\midrule
Xception & 0.0, 1.0 & -4.1328, 0.0032 & -3.5555, 0.0074 & -3.2377, 0.0119 & -3.9350, 0.0043 \\
Xception+Wavelet & 4.1328, 0.0032 & 0.0, 1.0 & 0.6890, 0.5102 & 0.3282, 0.7511 & -0.3569, 0.7303 \\
Xception+Wavelet+GWO & 3.5555, 0.0074 & -0.6890, 0.5102 & 0.0, 1.0 & -0.1031, 0.9204 & -0.7791, 0.4583 \\
Xception+Wavelet+Fox & 3.2377, 0.0119 & -0.3282, 0.7511 & 0.1031, 0.9204 & 0.0, 1.0 & -0.4959, 0.6332 \\
Xception+Wavelet+MGTO & 3.9350, 0.0043 & 0.3569, 0.7303 & 0.7791, 0.4583 & 0.4959, 0.6332 & 0.0, 1.0 \\
\bottomrule
\end{tabular}
}
\label{tab:xception_statistic17}
\end{table}

\begin{table}[h!]
\centering
\caption{Inception statistic and pvalue (ISIC-2017)}
\resizebox{\textwidth}{!}{%
\begin{tabular}{lccccc}
\toprule
\textbf{Model} & \textbf{Inception} & \textbf{Inception+Wavelet} & \textbf{Inception+Wavelet+IGWO} & \textbf{Inception+Wavelet+Fox} & \textbf{Inception+Wavelet+MGTO} \\
\midrule
Inception & 0.0, 1.0 & -1.9801, 0.0830 & -3.5697, 0.0072 & -3.3192, 0.0105 & -4.1243, 0.0033 \\
Inception+Wavelet & 1.9801, 0.0830 & 0.0, 1.0 & -2.5466, 0.0343 & -2.1246, 0.0663 & -3.5868, 0.0071 \\
Inception+Wavelet+IGWO & 3.5697, 0.0072 & 2.5466, 0.0343 & 0.0, 1.0 & 0.9300, 0.3795 & -1.3405, 0.2168 \\
Inception+Wavelet+Fox & 3.3192, 0.0105 & 2.1246, 0.0663 & -0.9300, 0.3795 & 0.0, 1.0 & -2.6962, 0.0272 \\
Inception+Wavelet+MGTO & 4.1243, 0.0033 & 3.5868, 0.0071 & 1.3405, 0.2168 & 2.6962, 0.0272 & 0.0, 1.0 \\
\bottomrule
\end{tabular}
}
\label{tab:inception_statistic17}
\end{table}

\begin{table}[h!]
\centering
\caption{DenseNet statistic and pvalue (ISIC-2017)}
\resizebox{\textwidth}{!}{%
\begin{tabular}{lccccc}
\toprule
\textbf{Model} & \textbf{DenseNet} & \textbf{DenseNet+Wavelet} & \textbf{DenseNet+Wavelet+IGWO} & \textbf{DenseNet+Wavelet+Fox} & \textbf{DenseNet+Wavelet+MGTO} \\
\midrule
DenseNet & 0.0, 1.0 & -0.0705, 0.9454 & -2.3359, 0.0477 & -1.0857, 0.3092 & -0.6534, 0.5318 \\
DenseNet+Wavelet & 0.0705, 0.9454 & 0.0, 1.0 & -1.1801, 0.2718 & -0.4831, 0.6419 & -0.3026, 0.7698 \\
DenseNet+Wavelet+IGWO & 2.3359, 0.0477 & 1.1801, 0.2718 & 0.0, 1.0 & 1.0702, 0.3157 & 1.2137, 0.2594 \\
DenseNet+Wavelet+Fox & 1.0857, 0.3092 & 0.4831, 0.6419 & -1.0702, 0.3157 & 0.0, 1.0 & 0.2389, 0.8171 \\
DenseNet+Wavelet+MGTO & 0.6534, 0.5318 & 0.3026, 0.7698 & -1.2137, 0.2594 & -0.2389, 0.8171 & 0.0, 1.0 \\
\bottomrule
\end{tabular}
}
\label{tab:densenet_statistic17}
\end{table}

\begin{table}[h!]
\centering
\caption{MobileNet statistic and pvalue (ISIC-2017)}
\resizebox{\textwidth}{!}{%
\begin{tabular}{lccccc}
\toprule
\textbf{Model} & \textbf{MobileNet} & \textbf{MobileNet+Wavelet} & \textbf{MobileNet+Wavelet+IGWO} & \textbf{MobileNet+Wavelet+Fox} & \textbf{MobileNet+Wavelet+MGTO} \\
\midrule
MobileNet & 0.0, 1.0 & -0.2828, 0.7844 & -3.3737, 0.0097 & -0.2502, 0.8087 & -0.6540, 0.5314 \\
MobileNet+Wavelet & 0.2828, 0.7844 & 0.0, 1.0 & -2.7292, 0.0258 & -0.0691, 0.9465 & -0.4482, 0.6658 \\
MobileNet+Wavelet+IGWO & 3.3737, 0.0097 & 2.7292, 0.0258 & 0.0, 1.0 & 1.3575, 0.2116 & 1.0879, 0.3082 \\
MobileNet+Wavelet+Fox & 0.2502, 0.8087 & 0.0691, 0.9465 & -1.3575, 0.2116 & 0.0, 1.0 & -0.2873, 0.7811 \\
MobileNet+Wavelet+MGTO & 0.6540, 0.5314 & 0.4482, 0.6658 & -1.0879, 0.3082 & 0.2873, 0.7811 & 0.0, 1.0 \\
\bottomrule
\end{tabular}
}
\label{tab:mobilenet_statistic17}
\end{table}

\begin{table}[h]
\centering
\caption{Hyperparameters of Swarm-Based Optimizers}
\label{tab:swarm_optimizers}
\begin{tabular}{|l|l|p{6.69cm}|c|c|}
\hline
\textbf{Optimizer} & \textbf{Hyperparameter} & \textbf{Description} & \textbf{Value} & \textbf{Range} \\
\hline
\multirow{2}{*}[+5pt]{MGTO} & \texttt{pp} & Probability of transition in the exploration phase & 0.03 & [0, 1] \\
\hline
\multirow{2}{*}{FOX} & \texttt{c1} & Coefficient of jumping (\(c_1\) in the paper) & 0.18 & [0, 1] \\
\cline{2-5}
 & \texttt{c2} & Coefficient of jumping (\(c_2\) in the paper) & 0.82 & [0, 1] \\
\hline
\multirow{2}{*}{IGWO} & \texttt{a\_min} & Lower bound of parameter \(a\) & 0.02 & [0, \(\infty\)) \\
\cline{2-5}
 & \texttt{a\_max} & Upper bound of parameter \(a\) & 2.2 & (0, \(\infty\)) \\
\hline
\end{tabular}
\end{table}

\begin{table}[h]
\centering
\caption{Hyperparameters of the Neural Network Model Optimized by Swarm Based Optimizers}
\label{tab:nn_hyperparameters}
\begin{tabular}{|l|p{6cm}|c|c|}
\hline
\textbf{Hyperparameter} & \textbf{Description} & \textbf{Lower Bound} & \textbf{Upper Bound} \\
\hline
\texttt{filters\_size} & Number of filters in the convolutional layer & 64 & 256 \\
\hline
\texttt{kernel\_size} & Size of the convolutional kernel & 3 & 9 \\
\hline
\texttt{lr} & Learning rate for the optimizer & \(1 \times 10^{-5}\) & \(1 \times 10^{-2}\) \\
\hline
\texttt{l2\_reg} & L2 regularization coefficient & \(1 \times 10^{-5}\) & \(1 \times 10^{-2}\) \\
\hline
\texttt{l1\_reg} & L1 regularization coefficient & \(1 \times 10^{-5}\) & \(1 \times 10^{-2}\) \\
\hline
\texttt{batch\_size} & Number of samples per gradient update & 16 & 128 \\
\hline
\texttt{epochs} & Number of training epochs & 10 & 100 \\
\hline
\texttt{att\_reg\_weight} & Regularization weight for the attention & \(1 \times 10^{-5}\) & \(1 \times 10^{-3}\) \\
\hline
\end{tabular}
\end{table}

Using a wavelet after DenseNet improves the model by combining its hierarchical feature extraction with the wavelet's ability to analyze features at multiple resolutions. The wavelet transform decomposes the DenseNet's features mentioned above into frequency bands, preserving high-value details like edges in the high-frequency part and larger patterns like lesion structure in the low-frequency part. Consequently, by removing the noise and less valuable features and emphasizing critical information by the wavelets output, the model's ability to detect localized and subtle lesion patterns, especially in imbalanced datasets, is improved\cite{zhou2019wavelet}. 

The Improved Grey Wolf Optimizer (IGWO) stands out for its ability to balance global exploration and local exploitation through adaptive coefficients. This property mainly benefits deep and interconnected models like Xception and DenseNet, which have highly non-convex loss surfaces. In DenseNet, where layers are densely connected, IGWO’s adaptive exploitation ensures effective utilization of the feature maps propagated through the network, resulting in improved precision and recall. Similarly, in Xception, which relies on depthwise separable convolutions, IGWO’s global search properties enhance the tuning of separable parameters, leading to superior performance.

The Fox Optimizer demonstrates its strength in lightweight models such as MobileNet. Its dynamic step size adjustment ensures precise updates in sparse parameter spaces, a critical feature for MobileNet’s depthwise separable convolutions. This property minimizes overfitting while maintaining high performance, which is especially useful in models with fewer parameters and less redundancy. By focusing on adaptive movement in the search space, the Fox Optimizer refines parameters effectively.

The modified Gorilla Troops Optimizer  (MGTO) combines gradient-based updates with adaptive learning rates, making it particularly suited for versatile architectures like DenseNet and Inception. MGTO’s use of momentum-like terms allows smooth optimization paths. In DenseNet, MGTO’s adaptive behavior complements the dense connectivity of layers, optimizing parameter updates in a way that fully leverages feature reuse across layers.

Our approach was evaluated by testing several models, including DenseNet, MobileNet, Xception, and Inception, on two datasets (ISIC-2016 and ISIC-2017). We focused on analyzing the impact of wavelet transformations and advanced optimizers on the models' performance.

Initially, we tested the models without an optimizer and wavelet to reach a baseline. \textbf{DenseNet} performed with an accuracy of 97.98\% on ISIC-2016 and 96.85\% on ISIC-2017.\textbf{Xception} and 
\textbf{MobileNet} and \textbf{Inception} performed slightly below DenseNet.
There is also the capacity for improvement, especially in recall and precision.

Using wavelet transformations significantly affected the model’s ability to extract image features, which increased the model’s accuracy in distinguishing between melanoma and non-melanoma features.

\textbf{Xception+Wavelet} showed a noticeable increase in accuracy, rising from 95.22\% to 97.94\% on ISIC-2016 and from 95.85\% to 96.35\% on ISIC-2017.
By combining wavelet transformations to focus on finer details, we see a noticeable improvement in all metrics.
Next, we added advanced optimizers to fine-tune the models. In this implementation, we used Grey Wolf Optimizer (GWO), Fox Optimizer, and Modified Gorilla Troops Optimizer (MGTO).

The Fox Optimizer showed its strength:
\textbf{MobileNet+Wavelet+Fox} reached an accuracy of 98.11\% on ISIC-2016 and 97.80\% on ISIC-2017, performing better than MobileNet alone.
\textbf{Inception+Wavelet+Fox} reduced false positives.

MGTO proved to be a robust choice:
\textbf{DenseNet+Wavelet+MGTO} achieved an impressive accuracy of 98.87\% on ISIC-2016 and 97.85\% on ISIC-2017.

\subsubsection*{Best Result}
According to the results in Tables 1-9, the DenseNet+Wavelet+Fox is the top performer for evaluation based on the ISIC-2016 datasets. The Fox optimizer's ability to reduce false positives by refining the model’s decision boundaries is a key factor. On the other hand, for models trained on the ISIC-2017 datasets, the Inception+Wavelet+MGTO is the best choice. MGTO's adaptability to the complex and diverse feature space of the datasets ensures a balanced and confident improvement in precision and recall. These combinations leverage the strengths of both wavelets and optimizers, resulting in models that are not only accurate but
also reliable for real-world melanoma classification tasks. This makes them ideal for clinical use, where accuracy and
trustworthiness are paramount.

\section*{Conlusion}
This article introduced a new method for detecting skin cancer, which combines optimization techniques with pre-trained networks and Wavelet transform to classify skin images effectively. The approach uses a convolutional neural network (CNN) enriched with Wavelet transforms to capture relevant image features. This architecture is then fine-tuned with its configuration and weight values using three advanced swarm-based optimization algorithms (MGTO-IGWO-FOX). These algorithms rely on a probability model to track the best solutions and parameters, boosting the algorithm's speed and accuracy in finding the optimal result. The proposed method was evaluated using the ISIC-2016 and ISIC-2017 datasets, and the results were compared with those of other methods. Based on the results, the combination of MGTO and FOX optimization and pre-trained networks with Wavelet Transform before the feature layer can increase the detection accuracy by at least 1.1\% compared to without the use of swarm-based optimizers in ISIC-2016 dataset and 2.05\% in ISIC-2017 dataset. Finally, the proposed SCD method can detect skin cancer disease, achieving accuracy of 98.11\% (in the ISIC-2016 dataset) and 97.95\% (in the ISIC-2017 dataset), at least 1\% higher than previous methods.

\section*{Data availability}
The datasets used in this study, ISIC 2016 and ISIC 2017, are publicly available and can be accessed at \href{https://challenge.isic-archive.com/data/}{https://challenge.isic-archive.com/data/} 

\section*{Code availability}
The source code for this study is openly available and can be accessed at \href{https://github.com/Parsa-Hatami/Enhancing-Skin-Cancer-Diagnosis-Using-Late-Discrete-Wavelet-Transform-and-New-Swarm-Based-Optimizers}{https://github.com/Parsa-Hatami/Enhancing-Skin-Cancer-Diagnosis-Using-Late-Discrete-Wavelet-Transform-and-New-Swarm-Based-Optimizers}

\end{document}